\documentclass[runningheads]{llncs}

\usepackage{eccv}
\usepackage{eccvabbrv}
\usepackage{graphicx}
\usepackage{booktabs}
\usepackage{adjustbox}
\usepackage[accsupp]{axessibility}
\usepackage{array}
\newcolumntype{P}[1]{>{\centering\arraybackslash}p{#1}}
\usepackage{hyperref}
\usepackage[dvipsnames]{xcolor}

\usepackage{colortbl}
\usepackage{tabularx}

\definecolor{tabfirst}{rgb}{1, 0.7, 0.7}

\definecolor{tabsecond}{rgb}{1, 0.85, 0.7}

\definecolor{tabthird}{rgb}{1, 1, 0.7}
\usepackage{multirow}
\usepackage {algpseudocode}
\usepackage{algorithmicx}
\usepackage{algcompatible}
\usepackage{longtable}
\begin{document}

\title{INTRA: Interaction Relationship-aware\\ Weakly Supervised Affordance Grounding}
\titlerunning{Interaction Relationship-aware Affordance Grounding}
\author{Ji Ha Jang\inst{1}$^{*}$ \and
Hoigi Seo\inst{1}$^{*}$ \and
Se Young Chun\inst{1,2}$^{\dagger}$}

\authorrunning{Jang \& Seo et al.}

\institute{$^1$Dept. of Electrical and Computer Engineering, $^2$INMC \& IPAI \\
Seoul National University, Republic of Korea\\
\email{\{jeeit17, seohoiki3215, sychun\}@snu.ac.kr}}
\maketitle
\def\thefootnote{*}\footnotetext{Authors contributed equally. $^{\dagger}$ Corresponding author.}

\begin{abstract}
Affordance denotes the potential interactions inherent in objects. The perception of affordance can enable intelligent agents to navigate and interact with new environments efficiently. Weakly supervised affordance grounding teaches agents the concept of affordance without costly pixel-level annotations, but with exocentric images. Although recent advances in weakly supervised affordance grounding yielded promising results, there remain challenges including the requirement for paired exocentric and egocentric image dataset, and the complexity in grounding diverse affordances for a single object. To address them, we propose  INTeraction Relationship-aware weakly supervised Affordance grounding (INTRA). Unlike prior arts, INTRA recasts this problem as representation learning to identify unique features of interactions through contrastive learning with exocentric images only, eliminating the need for paired datasets. Moreover, we leverage vision-language model embeddings for performing affordance grounding flexibly with any text, designing text-conditioned affordance map generation to reflect interaction relationship for contrastive learning and enhancing robustness with our text synonym augmentation. Our method outperformed prior arts on diverse datasets such as AGD20K, IIT-AFF, CAD and UMD. Additionally, experimental results demonstrate that our method has remarkable domain scalability for synthesized images / illustrations and is capable of performing affordance grounding for novel interactions and objects. Project page: \url{https://jeeit17.github.io/INTRA}
  \keywords{Affordance grounding \and Weak supervision \and Exocentric image \and Contrastive learning \and Interaction relation}
\end{abstract}

\begin{figure}[!t]
    \includegraphics[width=1\textwidth]{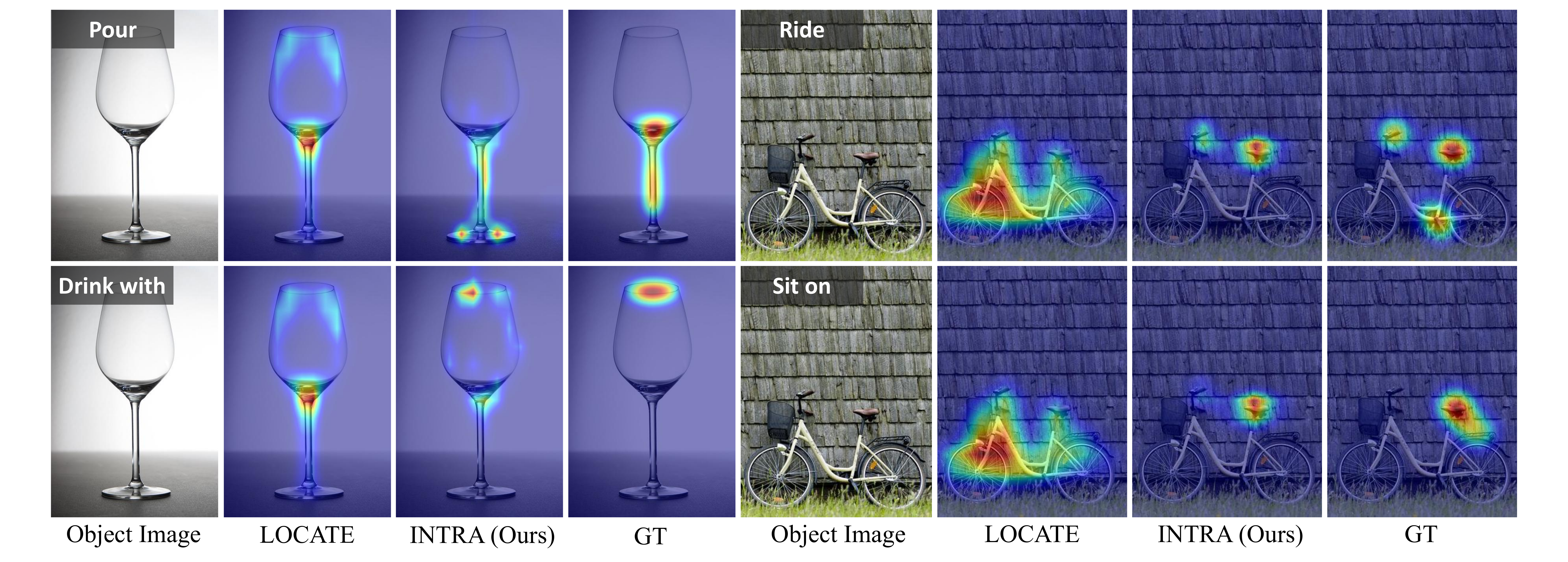}
    \caption{Prior works on weakly-supervised affordance grounding like LOCATE~\cite{li2023locate} often failed to ground different affordances for the same object. However, our proposed INTRA yielded finer and more accurate grounding results for them that are closer to the ground truth (GT) by considering interaction relationship among them.}
    \label{fig:fig2-1_weakly}
\end{figure}

\section{Introduction}
\label{sec:intro}
Affordance~\cite{gibson1986ecological} refers to the perceived possible interactions based on an object’s inherent or recognized properties (\textit{e.g.}, the rim of a wine glass affords sipping while stem of it affords holding). Humans can identify affordances of objects and interact with proper parts despite the diversity in their physical attributes. This ability can be acquired through individual learning, by directly interacting with objects, and observational learning~\cite{burke2010neural}, by observing others' interactions. The sense of affordance enables effective interaction in new environments or with novel objects, without step-by-step instructions~\cite{climbing}. Affordance plays an essential role across numerous applications involving intelligent agents, enabling them to provide flexible and timely responses in complex, dynamic environments~\cite{ardon2020affordances}. These applications include task planning, robot grasping, manipulation, scene understanding and action prediction~\cite{zhang2023affordancedriven,saycan2022arxiv, rana2023sayplan, ardon2019learning, geng2023rlafford, bahl2023affordances}. 

Affordance grounding is the task to teach intelligent systems how to locate possible action regions in objects for a certain interaction. While fully supervised learning~\cite{hadjivelichkov2023one, 9364360, amir2021deep,7759429} is the most straightforward approach, its reliance on costly annotations may limit its applicability across diverse contexts.
Another approach is weakly supervised learning, similar to human's observational learning~\cite{burke2010neural}, that does not require GT, but \emph{weak} labels. In this setting, \emph{exocentric} images illustrating human-object interactions, along with corresponding \emph{egocentric} images depicting the objects, are provided during training. During inference, intelligent systems perform affordance grounding on the egocentric images, identifying object parts relevant to the given interactions. Recent advances in weakly supervised affordance grounding~\cite{luo2022grounded, luo2022learning, li2023locate, nagarajan2019grounded} proposed to use \emph{pairs} of exocentric and egocentric images, yielding great performance.
The deep neural networks learn affordances by pulling features from exocentric and egocentric images closer, aiming to focus on object parts related to interactions.

However, weakly supervised affordance grounding remains challenging. Firstly, the requirement for current weak labels with \emph{pairs} of exocentric and egocentric images is still strong. Note that human observational learning does not usually require egocentric images. Secondly, a complex relationship between interactions exists, which has not been adequately addressed in prior works. Many instances in object-interaction relationships exhibit intricate many-to-many associations, occasionally with one entailing another. For example, some distinct interactions represent the same affordance regions (\textit{e.g.}, `wash' and `brush with' a tooth brush), and there are closely related interactions that always come together (\textit{e.g.}, `sip' usually includes `hold'. `ride' usually includes `sit on'). This complexity poses challenges in extracting interaction-relevant features based on image-level affordance labels, introducing biases towards objects in affordance grounding as illustrated in Fig.~\ref{fig:fig2-1_weakly} (LOCATE~\cite{li2023locate} often yielded similar affordance grounding with different interactions for the same object).

Here, we propose a novel weakly supervised affordance grounding method, INTRA (INTeraction Relationship-aware weakly supervised Affordance grounding) to address these unexplored challenges. While previous studies~\cite{luo2022grounded, nagarajan2019grounded} solved the weak supervision problem as supervised learning by pulling object features of exocentric and egocentric images closer and LOCATE~\cite{li2023locate} enhanced this approach by generating more localized pseudo labels based on prior information for exocentric images for supervised learning (\textit{i.e.}, containing human, object part, and background), our INTRA framework recasts the weak supervision problem as representation learning. This novel reformulation allows us to use \emph{weaker} labels (\textit{i.e.}, exocentric images only) for training so that the requirement to use \emph{pairs} of exocentric / egocentric images is now alleviated. Moreover, unlike prior works, our INTRA method actively exploits large language model (LLM) as well as the text encoder of the vision-language model (VLM) to leverage linguistic information and existing textual knowledge on affordances, which further enhances our interaction relationship-guided contrastive learning. This novel scheme also allows excellent scalability for unseen objects across diverse domains as well as zero-shot inference for novel interactions, which was not possible in prior arts. In summary, our main contributions are three-fold as follows:
\begin{itemize}
    \item {We propose a novel approach for weakly supervised affordance grounding by recasting the problem as representation learning and by leveraging VLM, leading to relaxing the need for paired training datasets for \emph{more} weak supervision and enhancing scalability across domains for unseen objects.}
    \item {We proposed INTRA, a novel method that consists of text synonym augmentation and text-conditioned affordance map generation module along with interaction relationship-guided contrastive learning, so that inference on unseen interactions is possible.}
    \item{Our INTRA outperforms the prior arts in weakly supervised affordance grounding on diverse datasets such as AGD20K, IIT-AFF, CAD and UMD, demonstrating both qualitative and quantitative excellence (see Fig.~\ref{fig:fig2-1_weakly}).}
\end{itemize}

\section{Related Works}
\subsection{Affordance Grounding}
\subsubsection{Supervised affordance grounding.}
Supervised affordance grounding methods~\cite{chen2023affordance, fang2018demo2vec} analyze interaction videos / images to predict affordance regions on an object, trained with pixel-level GT masks / heat maps. Though successful in localizing fine-grained affordance regions through supervised learning, they are limited by the costly GT mask annotation process and their limited generalizability to unseen objects. Furthermore, they require paired demonstration videos and target object images, making real-world application challenging.

\subsubsection{Weakly supervised affordance grounding.}
Weakly supervised affordance grounding methods~\cite{luo2022learning,luo2022grounded,nagarajan2019grounded,li2023locate, huang2018predicting, cornia2016deep, kummerer2016deepgaze, pan2017salgan, luo2023learning} offer the advantage of not requiring GT, but requiring weak labels such as exocentric images with interaction text labels. Prior works~\cite{luo2022learning,luo2022grounded,nagarajan2019grounded,li2023locate} mainly align interaction-relevant object features from both egocentric and exocentric images without considering the intrinsic properties of interactions. The framework in~\cite{nagarajan2019grounded} predicts object features engaged in interactions by analyzing human-object interaction videos. The works of~\cite{luo2022grounded, luo2022learning} preserve the correlation of affordance features from exocentric and egocentric images to learn affordances. The work of~\cite{li2023locate} enhances object feature extraction by adopting DINO-ViT~\cite{caron2021emerging} based Class Activation Maps (CAM)~\cite{zhou2016learning} and k-means clustering~\cite{macqueen1967some} for more explicit guidance. However, focusing solely on object features may introduce biases towards object, hindering the inference of multiple affordances for a single object. Our INTRA addresses this issue by considering the complex relationships between interactions using interaction relationship-guided contrastive loss, while ensuring the network remains attentive to the objects using object-variance mitigation loss.

\subsection{Foundation Models for Affordance Grounding}

\subsubsection{Self-supervised transformer.} 
Self-supervised transformers, extensively trained on large-scale datasets and scalability, possess robust representation power. Previous works~\cite{li2023locate, rashid2023language} have explored their potential in affordance grounding. DINO-ViT~\cite{caron2021emerging}, a vision transformer foundation model trained in a self-supervised manner, can identify both high-semantics such as overall information of the image and low-semantics such as details regarding specific object parts. This versatility has led advancements in various tasks, including classification, semantic segmentation~\cite{amir2021deep, li2023mask} and semantic correspondence~\cite{zhang2023tale}. LOCATE~\cite{li2023locate} leverages DINO-ViT to extract low-semantic information, resulting in performance improvements in affordance grounding. Our INTRA employed DINOv2~\cite{oquab2023dinov2} as an image encoder to extract information about objects and their constituent parts.

\subsubsection{Vision-language model.} 
The Vision-Language Model (VLM) is a class of models jointly pretrained on visual and language data for various downstream tasks~\cite{ning2023hoiclip, wan2024exploiting, yu2023zero}. VLM text encoders, trained through contrastive learning with image-text pairs, capture representations in the joint space of the images and text~\cite{radford2021learning, li2021align, li2022blip, li2023blip}. These text encoders, incorporating visual information, have demonstrated excellent performance across multiple tasks. ALBEF~\cite{li2021align} notably enhances vision and language representation learning by aligning image and text features before fusing them. While supervised affordance grounding methods leveraging VLM text encoders~\cite{nguyen2023open} have been explored, their application in weakly supervised affordance grounding remains underexplored. We propose a framework leveraging the text encoder of ALBEF to enable novel interactions, diverging from prior arts limited to inferring predetermined sets of affordances.

\subsubsection{Large language model.}
Understanding affordance relationships is crucial for affordance grounding, as it enables extending and linking learned visual cues, and reasoning about affordances for new objects, interactions, or situations. While prior works like~\cite{hou2021affordance} 
leverage semantically similar object properties and~\cite{luo2022learning} utilize affordance feature correlation, none directly exploit these relationships. We use these intricate relationships in affordance learning by adopting Large Language Models (LLMs). LLMs have gained prominence in robotics due to their profound natural language understanding, providing valuable priors about interactions and their complex relationships. Previous works~\cite{ahn2022can, singh2023progprompt, liang2023code, zhao2023chat} focus on extracting action knowledge, deriving task-specific plans, and grounding them in the physical world. LLMs have also been widely used in previous affordance studies~\cite{tang2023cotdet, mees2023grounding}, demonstrating their exceptional understanding of interactions.

\begin{figure}[!t]
     \includegraphics[width=\linewidth]{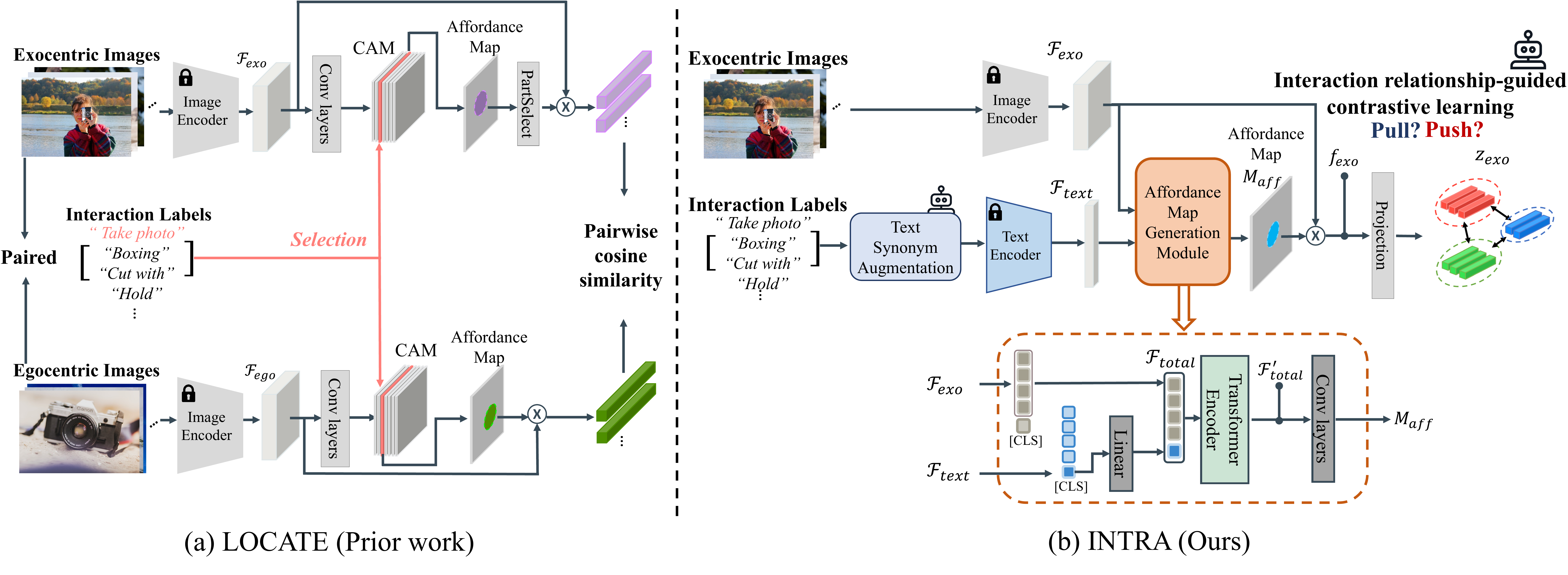}
    \caption{Overall frameworks of (a) LOCATE~\cite{li2023locate} and (b) INTRA (Ours). LOCATE takes paired exocentric and egocentric images to generate interaction-aware affordance maps (CAMs) for predefined interactions and then selects an interaction-related CAM by the given interaction label. In contrast, INTRA takes only exocentric images and interaction labels to yield an affordance map through our affordance map generation module. Training is done via interaction relationship-guided contrastive learning on exocentric features from affordance maps. Note that all encoder parameters are frozen.}
    \label{fig:fig2_method}
\end{figure}

\newcommand{\comb}[2]{{}_{#1}\mathrm{C}_{#2}}
\section{Method}
Prior arts in weakly supervised affordance grounding~\cite{luo2022learning,luo2022grounded,nagarajan2019grounded,li2023locate} typically align object features of paired exocentric (interaction with object) and egocentric (object only) images to learn interaction-related features. For example, as illustrated in Fig.~\ref{fig:fig2_method}(a), LOCATE~\cite{li2023locate} generates CAMs (affordance maps) from exocentric and egocentric images for a pre-determined interaction label, extracts egocentric feature as well as exocentric object parts feature selected by PartSelect module (pseudo label), and then trains the model by optimizing cosine similarity to align (pull) egocentric and exocentric object parts features. In contrast, we propose an alternative approach, INTRA, whose overall framework is illustrated in Fig.~\ref{fig:fig2_method}(b). Our text-conditioned affordance grounding framework of INTRA leverages VLM text encoder in affordance map generation module and employs text synonym augmentation to enhance robustness, as will be described in Sec.~\ref{text_condition}. Then, INTRA learn affordance grounding via our interaction relationship-guided contrastive learning, detailed in Sec.~\ref{contrast_learning}. The framework of INTRA as depicted in Fig.~\ref{fig:fig2_method}(b) clearly suggests two advantages over prior arts including LOCATE~\cite{li2023locate}: 1) it exploits exocentric images only and 2) INTRA admits novel interactions outside the pre-defined interaction set.

\subsection{Text-conditioned Affordance Grounding Framework} \label{text_condition}
To utilize the semantic meanings inherent in interaction labels and enable flexible inference on novel verbs, our text-conditioned affordance grounding framework generates affordance maps by conditioning image features with text features via our affordance map generation module where text and image features extracted from separately pre-trained text and image encoders are fused. In specific, as depicted in Fig.~\ref{fig:fig2_method}(b), deep features $\mathcal{F}_{exo} \in \mathbb{R}^{(h \times w) \times d}$ are obtained from the input exocentric images using DINOv2~\cite{oquab2023dinov2}, where $h$ and $w$ represent the height and width of the affordance map, and $d$ refers to the dimension of the feature. The text feature $\mathcal{F}_{text}$ of the given interaction is obtained using the ALBEF text encoder~\cite{li2021align}. See the supplementary material for further details on the rationale for employing DINOv2 and the ablation study on the text encoder.

\subsubsection{Affordance map generation module.}\label{map generation}
Before fusing text and image features, the class token of $\mathcal{F}_{text}$ passes through a single linear layer to align the separately pre-trained image and text embedding spaces and connect them, as shown to be effective in previous works~\cite{liu2024visual, zhu2023minigpt}. Subsequently, image features $\mathcal{F}_{exo}$ and the class token of text features are concatenated and processed through a transformer encoder for conditioning. The image feature part of the resulting vector is then projected using a multi-layered convolutional network and normalized using min-max normalization to obtain the affordance map $\mathcal{M}_{aff} \in \mathbb{R}^{h\times w}$. This affordance map represents the image regions in exocentric images most relevant to interactions. During inference, $\mathcal{M}_{aff}$  functions directly as an output heatmap, indicating the image regions in egocentric images most relevant to interactions.
\subsubsection{Text synonym augmentation.} \label{synonym_augment}
 To enhance the robustness of text conditioning, we integrate text synonym augmentation into our interaction embeddings. Initially, we generate $k_s$ synonyms for each interaction label using LLM. Subsequently, any synonyms overlapping with other interaction labels are removed. These synonyms are then randomly selected to substitue the text conditioning interaction embedding, while the original interaction label is retained for interaction relationship-guided contrastive learning. This module enhances overall performance by providing models with enriched interpretations of interactions.
 
\subsection{Interaction Relationship-guided Contrastive Learning} \label{contrast_learning}
Our INTRA learns via interaction relationship-guided contrastive learning by comparing exocentric image features across diverse interactions. 
Our contrastive learning consists of two key components, 1) extracting exocentric image features with affordance map and 2) designing loss for interaction relationship-guided contrastive learning, that enable the grounding of multiple affordances on a single object.

\subsubsection{Exocentric image feature extraction with affordance map.}
As described in Sec~\ref{map generation}, a text-conditioned affordance map, $\mathcal{M}_{aff}$, is generated to represent interaction-relevant image regions of exocentric images. Then, the exocentric image features $f_{exo}$ corresponding to the affordance map are extracted as follows: 
 \begin{equation}
    \centering
    f_{exo} = (1/hw){\Sigma_{i=1}^{h}\Sigma_{j=1}^{w}\mathcal{F}_{exo}(i,j) \cdot \mathcal{M}_{aff}(i, j)} \in \mathbb{R}^{d}.
    \label{eq:eq1_feature}
\end{equation}
The resulting $f_{exo}$ is then projected and normalized to obtain the exocentric image feature $z_{exo}$ using an MLP layer, which will be used for training. This projection layer was also used in previous works~\cite{chen2020simple, xue2024investigating, chen2021exploring}, which have demonstrated the necessity and efficiency of it.

\subsubsection{Loss design for interaction relationship-guided contrastive learning.}
Supervised contrastive learning~\cite{khosla2020supervised} effectively derives good representations for each class by focusing on common characteristics in positive pairs while disregarding those in negative pairs like other classes. However, in affordance grounding tasks, treating all other interaction classes as negative pairs may be inadequate due to the complex relationship among interactions. To mitigate this issue, we propose \emph{interaction relationship-guided contrastive loss}, $\mathcal{L}_{inter}$. Furthermore, considering the subtle meaning variations within single interaction classes depending on the object and context, we also propose \emph{object-variance mitigation loss}, $\mathcal{L}_{obj}$. Thus, the total loss for our INTRA is formulated as follows: 
\begin{equation}
    \centering
    \mathcal{L}_{total} = \mathcal{L}_{inter} + \lambda_{obj} \mathcal{L}_{obj}
    \label{eq:eq3_total_loss}
\end{equation}
 where $\lambda_{obj}$ denotes the control parameter of $\mathcal{L}_{obj}$.
\begin{figure*}[t!]
    \includegraphics[width=\linewidth]{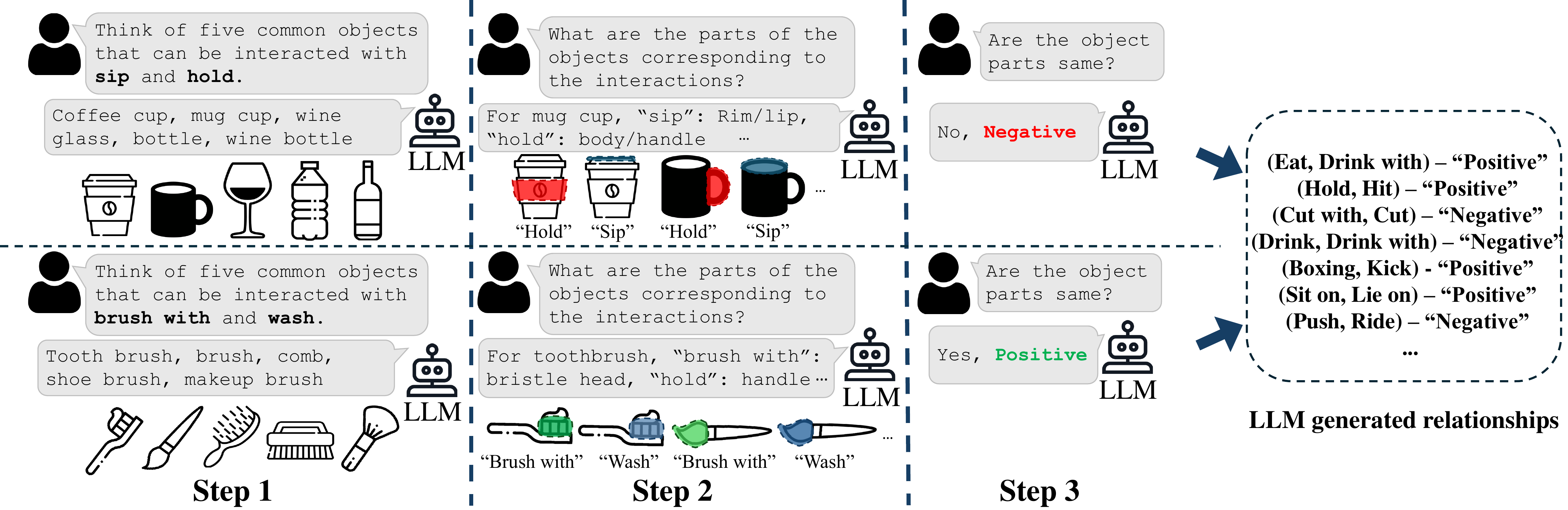}
    \caption{The overall scheme of interaction-relationship map ($\mathcal{R}$) generation. LLM classifies all pairs of interactions in the dataset as positive or negative through chain of thoughts. This process is based on reasoning if interactions occur on same object parts.}
    \label{fig:relationship}
\end{figure*}

\subsubsection{Interaction relationship-guided contrastive loss.}
In affordance grounding, treating all other interaction classes as negative pairs is inadequate due to the intricate relationships between interactions. For example, `Wash' and `Brush with' toothbrush or `Pour' and `Seal' bottle represent distinct interactions but act on the same object parts. Manually finding these relationships is time-consuming and impractical as the number of pairs grows quadratically with the number of interaction (see the supplementary). Moreover, although linguistic relationships like synonyms or co-occurrence were considered as substitutes, they are often inadequate and degrade performance. For example, `Sip' entails `Hold', but they act on different part of objects, and `Wash' and `Cut with' a knife have different meanings, but they act on the same blade. To mitigate this, we leverage LLM to determine if interaction pairs act on the same object part. Through Chain of Thoughts (CoT), interaction pairs are categorized as positive or negative in three steps as described in Fig.~\ref{fig:relationship}. In Step 1, LLM deduces five different objects where both interactions could be performed. In Step 2, LLM identifies object parts where these interactions could occur by considering five objects one by one, not simultaneously. In Step 3, if the identified parts of the interaction pair are the same, the pair is classified as positive; otherwise, negative. Positive pairs are assigned $1$ in the interaction-relationship map $\mathcal{R}$, and negative pairs are assigned $0$. We propose interaction-relationship guided contrastive loss by integrating $\mathcal{R}$ into supervised contrastive learning as follows:
\begin{equation} 
\small
    \mathcal{L}_{inter} = \sum\limits_{i=1}^{2N} \frac{-1}{2N_{y_{i}}-1}\sum\limits_{j=1}^{2N}\mathcal{R}_{(y_i, y_j)} \cdot 
    \log{\frac{\exp{(z_{exo}^i \cdot z_{exo}^j / \tau)}}{\sum\limits_{k=1}^{2N} \mathbf{1}_{i \neq k} \cdot \exp{(z_{exo}^i \cdot z_{exo}^k / \tau)}}}
    \label{eq:eq2_interaction_loss}
\end{equation}
where $i$, $j$ are sample indices, $y_i$, $y_j$ are class labels, $N_{y_{i}}$ is the number of samples in the batch labeled with $y_i$, $N$ is the total number of distinct samples in the batch, $z_{exo}^j$ is the exocentric image feature vector of each sample, $\tau$ is the temperature, and $\mathcal{R}_{(y_i, y_j)}$ is the value of $(y_i, y_j)$ pair in interaction-relationship map.

\subsubsection{Object-variance mitigation loss.}
 In the context of affordance, the interpretation of the same interaction can vary significantly based on the object and context. For instance, `Hold' a baseball bat and a cup may seem similar since both involve grasping an object. However, the former involves gripping the bat's slender part, while the latter entails holding the cup's rounded, protruding part. To address this variance within the same interaction category, we implemented an object-variance mitigation loss $\mathcal{L}_{obj}$ as follows:
\begin{equation} 
\small
    \sum\limits_{i=1}^{2N} \frac{-1}{2N_{o_{i}}-1}\sum\limits_{j=1}^{2N}\mathbf{1}_{o_{i}=o_{j}} \cdot \log{\frac{\exp{(z_{exo}^i \cdot z_{exo}^j / \tau)}}{\sum\limits_{k=1}^{2N} \mathbf{1}_{i \neq k} \cdot \exp{(z_{exo}^i \cdot z_{exo}^k / \tau)}}}
    \label{eq:eq3_obj_loss}
\end{equation}
where $o_i$, $o_j$ denote object class of $i$ and $j$.

\section{Experiments}
\subsection{Experimental Setting}
\subsubsection{Dataset and metrics.}
We conducted an evaluation of our method using the Affordance Grounding Dataset (AGD20K)~\cite{luo2022learning}. AGD20K comprises both exocentric and egocentric images, with 20,061 exocentric images and 3,755 egocentric images labeled with 36 affordances. The dataset support evaluation under two settings: 1) the `Seen' setting, where the object categories of the training and testing sets are identical, and 2) the `Unseen' setting, where no objects overlap between the training and test sets. Our approach only used exocentric images in training for all experiments, while other approaches were trained using both egocentric and exocentric images. We employed three evaluation metrics commonly employed in previous affordance grounding methodologies: 1) Kullback-Leibler Divergence (KLD), 2) Similarity (SIM), 3) and Normalized Scanpath Saliency (NSS). These metrics were utilized to quantify the similarity between the distributions of ground truth heatmaps and predicted affordance grounding.

\subsubsection{Implementation details.}
We employed DINOv2 as the image encoder and ALBEF, fine-tuned with RefCOCO+, as the text encoder. ChatGPT-4~\cite{achiam2023gpt} served as the LLM. Images were resized to 384$\times$384, then cropped to 336$\times$336. Training utilized the Adam optimizer~\cite{kingma2014adam} with a learning rate of 2e-4 and a batch size of 256. The hyperparameter $\lambda_{obj}$ was set to 4, and all experiments were conducted on a single NVIDIA A100 GPU. More details are provided in the supplementary.

\begin{table}[!t]
\caption{Quantitative results of ours and other baselines~\cite{pan2021unveiling, gao2021ts, mai2020erasing, nagarajan2019grounded, luo2022learning, luo2022grounded, li2023locate} on the AGD20K dataset. $\uparrow$ / $\downarrow$ indicates that higher / lower the metric is, the better the model performs. INTRA outperformed all baselines, despite being trained only with exocentric images, whereas other models incorporated both exocentric and egocentric images during training.}
\resizebox{\textwidth}{!}{%
\begin{tabular}{cclcccccc}
\hline
\multicolumn{3}{c}{\multirow{2}{*}{Prior works}}                                                                                              & \multicolumn{3}{c}{Seen} & \multicolumn{3}{c}{Unseen} \\
\multicolumn{3}{c}{}                                                                                                                          & mKLD$\downarrow$   & mSIM$\uparrow$   & mNSS$\uparrow$   & mKLD$\downarrow$    & mSIM$\uparrow$    & mNSS$\uparrow$   \\ \hline
\multicolumn{2}{c}{\multirow{3}{*}{\begin{tabular}[c]{@{}c@{}}Weakly Supervised\\ Object Localization\end{tabular}}}         & EIL~\cite{mai2020erasing}            & 1.931  & 0.285  & 0.522  & 2.167   & 0.227   & 0.330  \\
\multicolumn{2}{c}{}                                                                                                         & SPA~\cite{pan2021unveiling}            & 5.528  & 0.221  & 0.357  & 7.425   & 0.167   & 0.262  \\
\multicolumn{2}{c}{}                                                                                                         & TS-CAM~\cite{gao2021ts}         & 1.842  & 0.260  & 0.336  & 2.104   & 0.201   & 0.151  \\ \hline
\multirow{5}{*}{\begin{tabular}[c]{@{}c@{}}Weakly Supervised\\ Affordance Grounding\end{tabular}} & \multirow{4}{*}{Exo+Ego} & Hotspots~\cite{nagarajan2019grounded}       & 1.773  & 0.278  & 0.615  & 1.994   & 0.237   & 0.557  \\
   &                          & Cross-view-AG~\cite{luo2022learning}  & 1.538  & 0.334  & 0.927  & 1.787   & \cellcolor{tabthird}0.285   & 0.829  \\
   &                          & Cross-view-AG+~\cite{luo2022grounded} & \cellcolor{tabthird}1.489  & \cellcolor{tabthird}0.342  & \cellcolor{tabthird}0.981  & \cellcolor{tabthird}1.765   & 0.279   & \cellcolor{tabthird}0.882  \\
   &                          & LOCATE~\cite{li2023locate}         & \cellcolor{tabsecond}1.226  & \cellcolor{tabsecond}0.401  & \cellcolor{tabsecond}1.177  & \cellcolor{tabsecond}1.405   & \cellcolor{tabsecond}0.372   & \cellcolor{tabsecond}1.157  \\ \cline{2-9} 
   & Exo                      & \textbf{INTRA (Ours)}    & \cellcolor{tabfirst}1.199  & \cellcolor{tabfirst}0.407  & \cellcolor{tabfirst}1.239  & \cellcolor{tabfirst}1.365   & \cellcolor{tabfirst}0.375   & \cellcolor{tabfirst}1.209  \\ \hline
\end{tabular}%
}
\label{tab:tab1_quant}
\end{table}
\begin{table}[!t]
\caption{Quantitative results on the modified IIT-AFF, CAD, and UMD dataset for our method and other baselines~\cite{luo2022grounded, luo2022learning, li2023locate}. Models were trained in the `Seen' setting of AGD20K and tested on the datasets without additional training. INTRA outperformed all baselines on all metrics across all datasets. * Objects with affordances that prior works are unable to predict were eliminated from the datasets for fairness, wheares our method can infer affordances on novel interactions.}
\resizebox{\textwidth}{!}{
\begin{tabular}{lccccccccc}
\hline
            & \multicolumn{3}{c}{IIT-AFF*~\cite{nguyen2017object}} & \multicolumn{3}{c}{CAD*~\cite{sawatzky2017weakly}} & \multicolumn{3}{c}{UMD*~\cite{myers2015affordance}} \\
            & mKLD$\downarrow$         & mSIM$\uparrow$         & mNSS$\uparrow$         & mKLD$\downarrow$        & mSIM$\uparrow$        & mNSS$\uparrow$        & mKLD$\downarrow$        & mSIM$\uparrow$        & mNSS$\uparrow$        \\ \hline
Cross-View-AG~\cite{luo2022learning}      & \cellcolor{tabthird}3.856       & \cellcolor{tabthird}0.096        & 0.849        & 2.568       & 0.173       & 0.589       & \cellcolor{tabthird}4.721      & \cellcolor{tabthird}0.014       & \cellcolor{tabthird}1.287       \\
Cross-View-AG+~\cite{luo2022grounded}     &  3.920       & 0.095        & \cellcolor{tabthird}1.072         & \cellcolor{tabthird}2.529       & \cellcolor{tabthird}0.176       &  \cellcolor{tabsecond}0.663       & 4.753      & 0.013       & 1.227       \\
LOCATE\cite{li2023locate}      & \cellcolor{tabsecond}3.315        &  \cellcolor{tabsecond}0.115        &  \cellcolor{tabsecond}1.709        &  \cellcolor{tabsecond}2.528      &  \cellcolor{tabsecond}0.187       & \cellcolor{tabthird}0.558       &  \cellcolor{tabsecond}4.083       &  \cellcolor{tabsecond}0.026       &  \cellcolor{tabsecond}2.699       \\
\textbf{INTRA(Ours)} & \cellcolor{tabfirst}2.663       & \cellcolor{tabfirst}0.148        & \cellcolor{tabfirst}2.511        & \cellcolor{tabfirst}2.095       & \cellcolor{tabfirst}0.243       & \cellcolor{tabfirst}1.259       & \cellcolor{tabfirst}3.081       & \cellcolor{tabfirst}0.062       & \cellcolor{tabfirst}4.195       \\ \hline
\end{tabular}
}
\label{tab:tab3_additional}
\end{table}

\subsection{Comparison to State-of-the-art Methods}
To comprehensively assess our method, we conduct quantitative and qualitative comparisons with state-of-the-art weakly-supervised grounding methods, incorporating a user study. We further expand our experiments to include additional datasets~\cite{nguyen2017object, sawatzky2017weakly, myers2015affordance} for a comprehensive evaluation. Refer to the supplementary materials for more details on the experimental settings.

\subsubsection{Quantitative results.}
We evaluated previous works~\cite{luo2022learning, luo2022grounded, li2023locate, pan2021unveiling, nagarajan2019grounded, mai2020erasing, gao2021ts} and our method based on the metrics mentioned above. Tab.~\ref{tab:tab1_quant} shows the quantitative comparison results of our method with prior arts. In both `Seen' and `Unseen' setting, our approach surpasses the baseline performances across all three metrics: KLD, SIM, and NSS, thereby setting a new state-of-the-art.

\subsubsection{Results on additional datasets.}
We evaluated the generalization and robustness of the INTRA framework, along with previous works~\cite{luo2022learning,luo2022grounded, li2023locate} trained in the `Seen' setting of AGD20K, on the IIT-AFF~\cite{nguyen2017object}, CAD~\cite{sawatzky2017weakly}, and UMD~\cite{myers2015affordance} datasets. The experiment was conducted in the `Seen' setting due to overlapping objects between these datasets and AGD20K. Each GT was processed in the same way as when evaluating the AGD20K test set. Despite significant domain gaps across datasets, INTRA outperformed in all metrics on all datasets, demonstrating its superior generalizability as shown in Tab.~\ref{tab:tab3_additional}. Further details of the experiment can be found in the supplementary material.

\begin{figure*}
    \includegraphics[width=\linewidth]{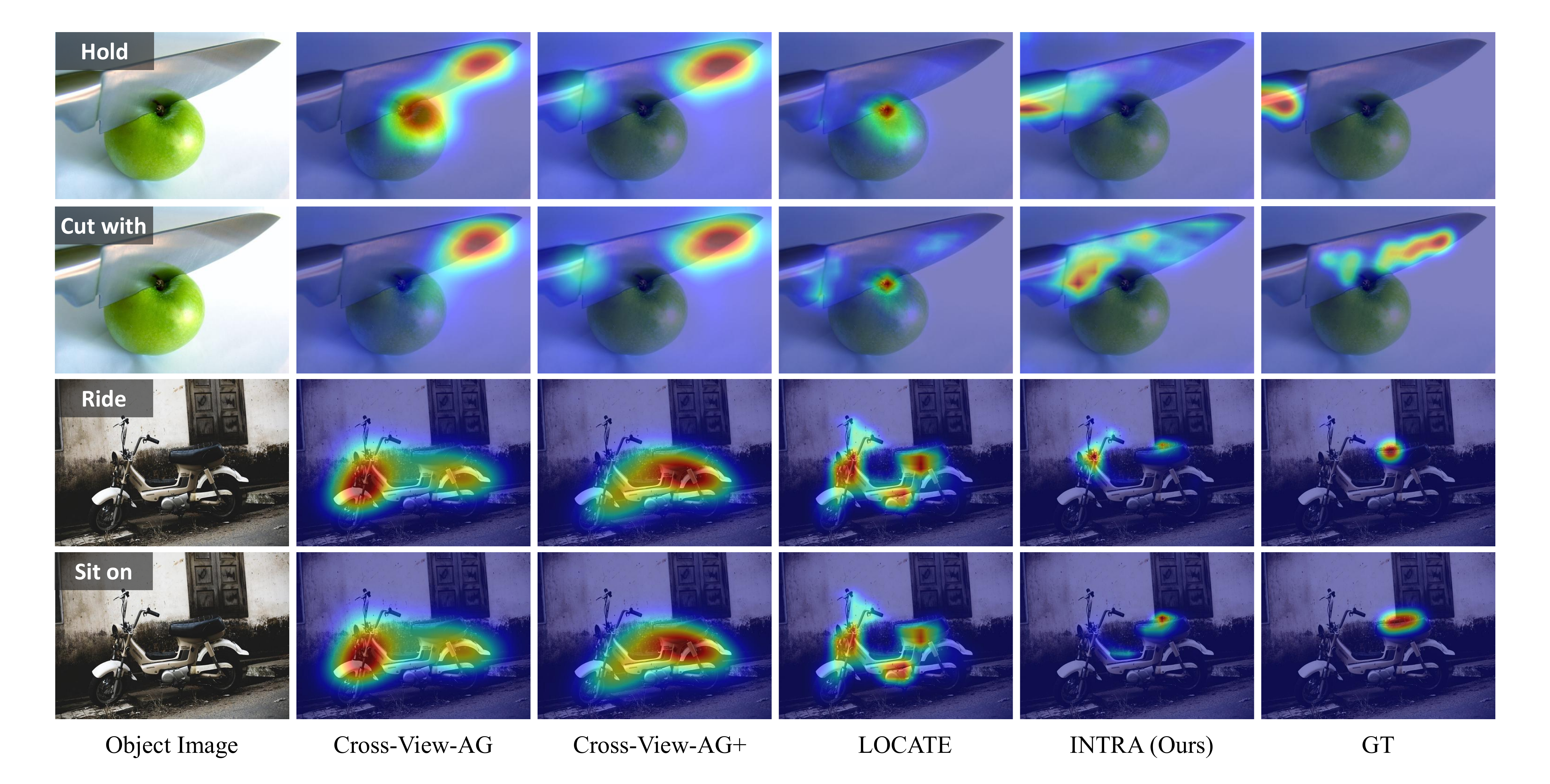}
    \caption{Qualitative results of INTRA (Ours) and baseline models~\cite{luo2022grounded, luo2022learning, li2023locate} on grounding affordances of multiple potential interactions on a single object. INTRA precisely localizes relevant interaction spots for each interaction. For example, with a knife, it grounds the handle for `Hold' and the blade for `Cut with'. For a motorcycle, it accurately grounds the saddle for `Sit on'. Additionally, for `Ride', it grounds both the handle and saddle, slightly deviating from the GT but still producing reasonable results, as we usually interacts with handle and saddle to `Ride' a motorcycle.}
    \label{fig:fig5_seperability}
    
    \includegraphics[width=\linewidth]{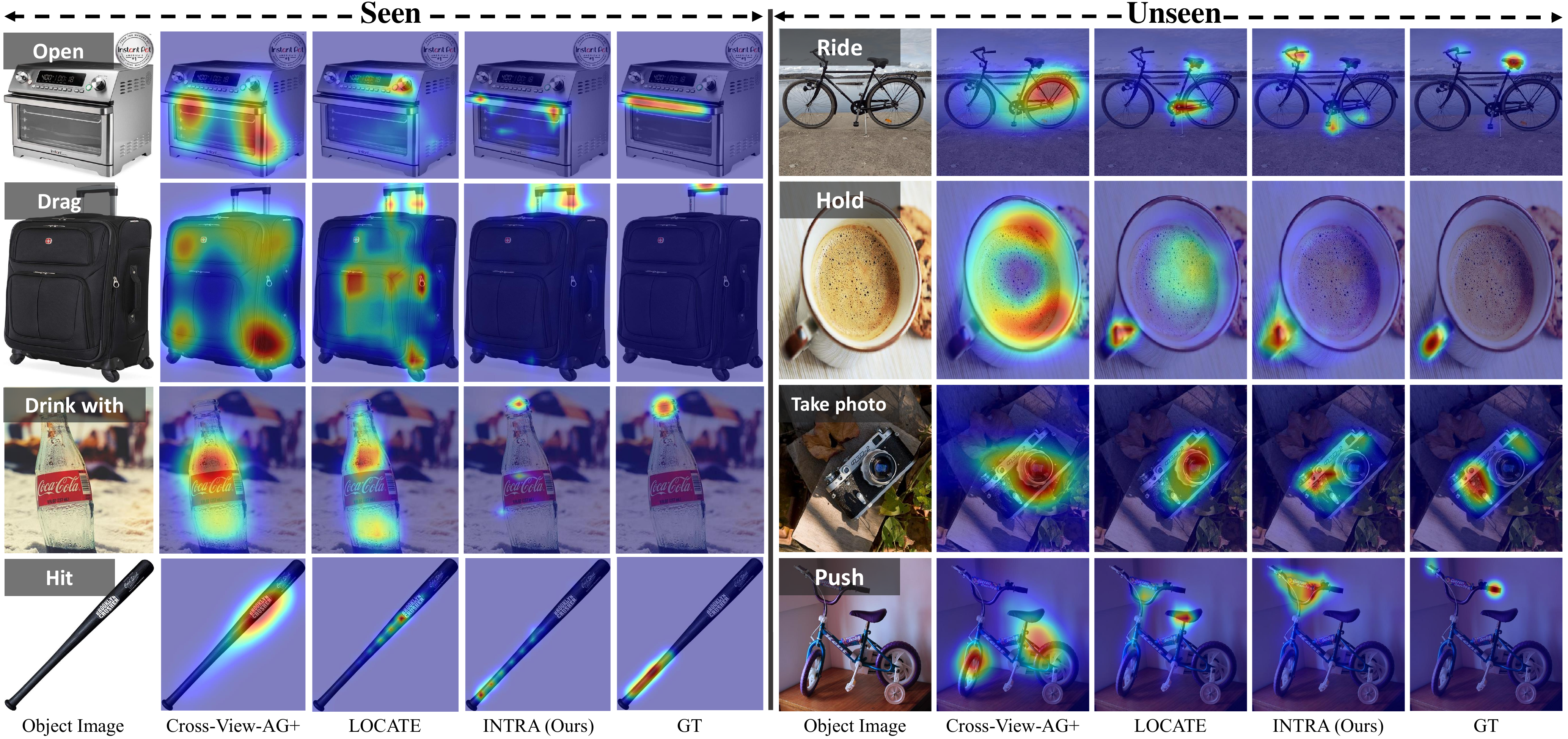}
    \caption{Qualitative results comparison between our approach and other baselines~\cite{luo2022grounded, luo2022learning, li2023locate}. Our approach, INTRA, demonstrates superior precision and detail in grounding affordances compared to the baselines. For instance, in the example of `Drag', while baselines either fail to localize the handle or erroneously ground several other parts, INTRA accurately identifies and grounds the handle of a suitcase with finesse.} 
    \label{fig:fig6_quali}
\end{figure*}

\subsubsection{Qualitative results.}
Fig.~\ref{fig:fig5_seperability} and Fig.~\ref{fig:fig6_quali} show our superior grounding precision compared to the baselines, being closer to the GT and finer in granularity. INTRA precisely identifies the exact object part for a given affordance, unlike the baselines, which ground the same parts regardless of the affordances provided.
\begin{table}[t!]
\caption{The result of user study on validity, finesse, and separability. Users were asked to score a 5-point scale, and we averaged it for mean opinion score (MOS).
}
\centering
\begin{tabular}{p{3.5cm} P{2cm} P{2cm} P{2cm}}
\hline
               & Validity & Finesse & Separability \\ \hline
Cross-View-AG+~\cite{luo2022grounded} & 2.897    & \cellcolor{tabthird}3.022   & \cellcolor{tabthird}2.732        \\
LOCATE~\cite{li2023locate}         & \cellcolor{tabsecond}3.054    & 2.573   & 2.651        \\
\textbf{INTRA (Ours)}    & \cellcolor{tabfirst}3.134    & \cellcolor{tabsecond}3.112   & \cellcolor{tabfirst}3.221        \\ 
Ground Truth   & \cellcolor{tabthird}2.905    & \cellcolor{tabfirst}3.334   & \cellcolor{tabsecond}3.160        \\ \hline
\end{tabular}

\label{tab:tab2_user}
\end{table}
\begin{table}[tp!]
\begin{minipage}[t]{0.49\textwidth}
\captionof{table}{Quantitative results of ablation study on our loss design. We incrementally added each component of the losses to examine their impact.
} 
\begin{adjustbox}{width=\linewidth}
\begin{tabular}{lcccccc}
\hline
\multicolumn{1}{c}{}                        & \multicolumn{3}{c}{Seen}                                                          & \multicolumn{3}{c}{Unseen}                                                        \\ \cline{2-7} 
\multicolumn{1}{c}{}                        & \multicolumn{1}{c}{mKLD$\downarrow$}  & \multicolumn{1}{c}{mSIM$\uparrow$}  & \multicolumn{1}{c}{mNSS$\uparrow$}  & \multicolumn{1}{c}{mKLD$\downarrow$}  & \multicolumn{1}{c}{mSIM$\uparrow$}  & \multicolumn{1}{c}{mNSS$\uparrow$}  \\ \hline
\multicolumn{1}{l}{baseline}                & \multicolumn{1}{c}{1.678} & \multicolumn{1}{c}{0.338} & \multicolumn{1}{c}{0.891} & \multicolumn{1}{c}{1.581} & \multicolumn{1}{c}{0.300} & \multicolumn{1}{c}{1.100} \\
$\mathcal{L}_{inter}$                      & 1.439                     & 0.334                     & 1.031                     & 1.569                     & 0.292                     & 1.133                     \\
$\mathcal{L}_{obj}$                        & 1.336                     & 0.387                     & 1.218                     & 1.521                     & 0.334                     & 1.042                     \\
$\mathcal{L}_{inter}$+$\mathcal{L}_{obj}$ & 1.199                     & 0.407                     & 1.239                     & 1.365                     & 0.375                     & 1.209                     \\ \hline
\end{tabular}
\end{adjustbox}
\label{tab:tab_loss}
\end{minipage}
\begin{minipage}[t]{0.49\textwidth}
\captionof{table}{Quantitative results of ablation study on different $\mathcal{R}$.  $\mathcal{L}_{WordNet}$, $\mathcal{L}_{Word2Vec}$ are calculated using word similarity from WordNet~\cite{miller1995wordnet}, Word2Vec~\cite{mikolov2013efficient}, respectively. $\mathcal{L}_{Co-occur.}$ used co-occurrence probability in GloVe~\cite{pennington2014glove}.} 
\begin{adjustbox}{width=\linewidth}
\begin{tabular}{lcccccc}
\hline
          & \multicolumn{3}{c}{Seen}                                                                  & \multicolumn{3}{c}{Unseen}                                                                \\ \cline{2-7}
          & mKLD$\downarrow$ & mSIM$\uparrow$ & mNSS$\uparrow$ & mKLD$\downarrow$ & mSIM$\uparrow$ & mNSS$\uparrow$ \\ \hline
$\mathcal{L}_{WordNet}$ & 1.701                         & 0.282                       & 0.710                       & 1.698                         & 0.277                       & 0.937                       \\
$\mathcal{L}_{Co-occur.}$ & 1.519                         & 0.309                       & 0.988                       & 1.639                         & 0.274                       & 1.101                       \\
$\mathcal{L}_{Word2Vec}$ & 1.547                         & 0.302                       & 0.958                       & 1.679                         & 0.270                       & 0.980                       \\
$\mathcal{L}_{inter}$\textbf{(Ours)} & \textbf{1.439}                         & \textbf{0.334}                       & \textbf{1.031}                       & \textbf{1.569}                         & \textbf{0.292}                       & \textbf{1.133}                       \\
\hline
\end{tabular}
\end{adjustbox}

    \label{tab:tab_inter}

\end{minipage}

 \end{table}

\subsubsection{User study.}
Affordance grounding can be ambiguous depending on context and interpretation, thus relying solely on metrics for evaluation has limitations. Hence, we conducted a user study comparing Cross-View-AG+~\cite{luo2022grounded}, LOCATE~\cite{li2023locate}, GT, and INTRA (Ours) across three categories: 1) Validity: assessing heatmap reasonableness, 2) Finesse: measuring heatmap detail, 3) Separability: determining the accuracy of the heatmap when different affordances are assigned to the same object. A total of 936 responses were collected for randomly selected samples from 104 respondents. Results presented in Tab.~\ref{tab:tab2_user} demonstrate that our approach outperforms baselines and par on GT based on human perception.

\subsection{Ablation Studies}
We validate our pipeline design choices and parameters with ablation studies. This section includes ablation studies on loss design, adoption of LLM, and text synonym augmentation. Refer to the supplementary for further ablation studies.
\subsubsection{Ablation study on loss design.}
To assess the individual impact of the components comprising loss on its overall performance, we analyzed by incrementally adding components. We started with the most basic element: a normal supervised contrastive loss. Subsequently, we sequentially added an interaction relationship-guided loss and an object-variance mitigation loss. The performance outcomes of these incremental modifications were thoroughly evaluated to understand their contributions, as represented in the Tab.~\ref{tab:tab_loss}. 

\subsubsection{Ablation study on adoption of LLM.}
Adopting LLM to create the relationship map was essential given the intricate nature of affordances. We experimented with various methods to create the relationship map to validate this choice. We measured the similarity of interaction pairs using WordNet~\cite{miller1995wordnet} and Word2Vec~\cite{mikolov2013efficient}, or computed co-occurence probability of interaction pairs with Glove~\cite{pennington2014glove}. Based on these measurements, we created an Interaction-relationship Map and trained the INTRA framework. The results are in the Tab.~\ref{tab:tab_inter}. 

\subsubsection{Ablation study on text synonym augmentation.} We conducted an ablation study on the effectiveness of text synonym augmentation on overall performance. We compared performance with and without the module. The module improved performance by up to 21.93\%, particularly in the `Unseen' setting, enriching models with varied meanings of interactions. Additionally, to test its effectiveness on novel verb inference, we deliberately omitted the subset `Hold' (24.17\% of training data) and then performed inference on `Hold'. The module boosted performance for novel verbs by up to 58.06\%. Similar tendencies were observed for other verbs. Detailed results are available in the supplementary.
\begin{figure*}[t]
    \includegraphics[width=\linewidth]{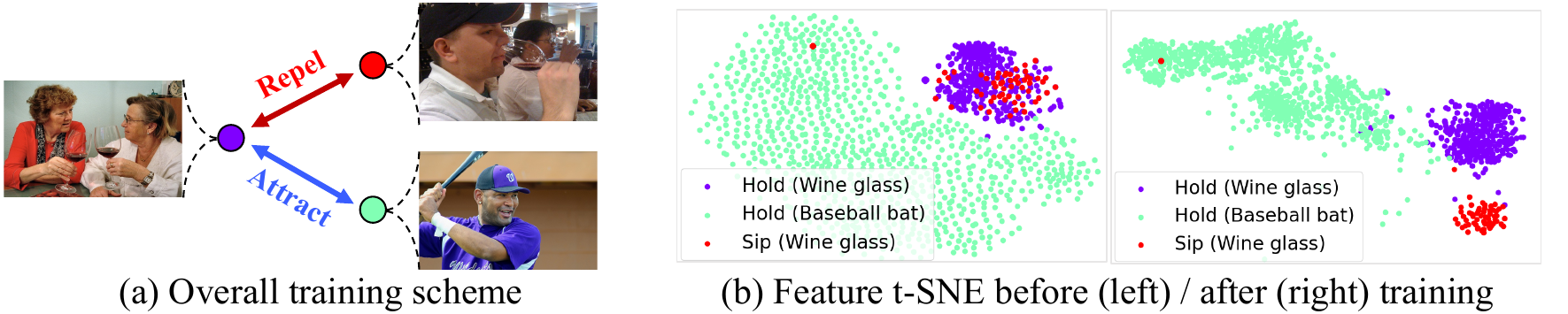}
    \caption{An illustration of interaction relationship-guided contrastive learning and t-SNE~\cite{van2008visualizing} visualization of feature distribution. (a) In interaction relationship-guided contrastive learning, positive interaction pairs attract each other, while others repel. (b) t-SNE visualization of DINOv2~\cite{oquab2023dinov2} class token and $f_{exo}$ from INTRA, showing that features of positive interaction pairs become closer as learning progresses.}
    \label{fig:conjecture}
\end{figure*}

\section{Discussion}
\subsection{Effect of Interaction Relationship-guided Contrastive Loss}
Our rationale for learning affordance grounding solely with exocentric images relies on the consistent presence of humans within these images. By repelling common features of negative pairs, such as human parts, the images effectively exclude irrelevant elements. Conversely, positive pairs, sharing the desired feature of the object—specifically, the rim of the object near the face—facilitate learning by attracting these relevant features (see Fig.~\ref{fig:conjecture}(a)). To visualize the effectiveness of our loss in learning interaction-relevant features in similar images, we examine the feature distributions of `Hold' and `Sip' a wine glass,  involving distinct affordances. Prior to training, these distributions overlap. However, after training with our loss function, the feature distribution for `Hold wine glass' aligns more closely with `Hold baseball bat' than with `Sip wine glass'. This indicates that our loss function effectively discriminates between the characteristics of different interactions without exhibiting bias towards objects (see Fig.~\ref{fig:conjecture}(b)). Detailed explanation is illustrated in the supplementary.

\subsection{Feasibility Study on Generalization Property of INTRA}
\begin{figure}[!t]
    \includegraphics[width=\linewidth]{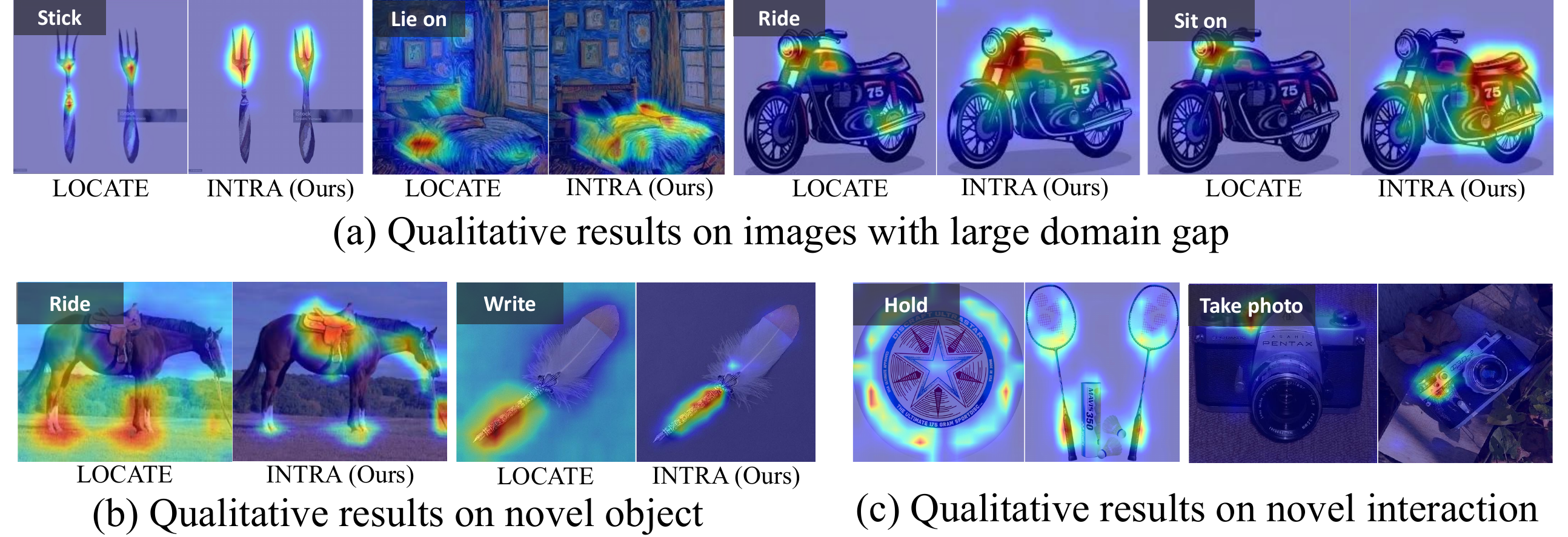}
    \caption{
Qualitative results of feasibility study: (a) Inference on diverse images with significant domain gap such as pixel arts and paintings. (b) Inference on novel objects that were not in the training data. (c) Inference on unseen novel interactions. INTRA demonstrates superior grounding accuracy in (a)-(c) compared to LOCATE~\cite{li2023locate}, showing proper affordance region inference without explicit training.}
\label{fig:fig_discussion}
\end{figure}

INTRA excels in affordance grounding on images with large domain gaps, such as pixel art and paintings, as illustrated in Fig.~\ref{fig:fig_discussion}(a). Furthermore, our method showcases strong generalization abilities for novel objects like a horse and quill, not present in the training set, as shown in Fig.~\ref{fig:fig_discussion}(b). Additionally, despite deliberately not being trained on specific interaction classes like `Hold' and `Take photo' for experiment, INTRA successfully infers their affordances, as depicted in Fig.~\ref{fig:fig_discussion}(c). More results and detailed experimental settings are in the supplementary. One possible explanation for this generalization property is that our INTRA employs VLM so that diverse domains and novel object can be dealt with without explicitly tuning for them. Another explanation is INTRA's contrastive training that may achieve better representation learning.

\section{Conclusion}
In this paper, we introduce  INTRA, a novel framework reformulating the weakly supervised affordance grounding with representation learning. We suggest interaction relationship-guided contrastive learning, informed by affordance knowledge from LLM. Furthermore, INTRA actively leverages VLM text embedding in proposed text-conditioned affordance map generation for flexible affordance grounding, further bolstered by text synonym augmentation for robustness. INTRA achieves state-of-the-art performance across diverse datasets, relying solely on exocentric images for training, unlike prior methods that also use egocentric images. Moreover, our method demonstrates generalization feasibility on novel objects, interactions, and images with significant domain gaps.

\section*{Acknowledgements}
This work was supported by Institute of Information \& communications Technology Planning \& Evaluation (IITP) grant funded by the Korea government(MSIT) [NO.RS-2021-II211343, Artificial Intelligence Graduate School Program (Seoul National University)], the National Research Foundation of Korea(NRF) grant funded by the Korea government(MSIT) (No. NRF-2022M3C1A309202211) and AI-Bio Research Grant through Seoul National University. Also, the authors acknowledged the financial support from the BK21 FOUR program of the Education and Research Program for Future ICT Pioneers, Seoul National University.

\bibliographystyle{splncs04}
\bibliography{main}

\renewcommand{\thefigure}{S\arabic{figure}}
\renewcommand{\thetable}{S\arabic{table}}
\renewcommand{\thesection}{S\arabic{section}}

\title{Supplementary Material for \\INTRA: Interaction Relationship-aware\\ Weakly Supervised Affordance Grounding}
\titlerunning{Interaction Relationship-aware Affordance Grounding (Supp.)}

\author{Ji Ha Jang\inst{1}$^{*}$ \and
Hoigi Seo\inst{1}$^{*}$ \and
Se Young Chun\inst{1,2}$^{\dagger}$}

\authorrunning{Jang \& Seo et al.}
\institute{$^1$Dept. of Electrical and Computer Engineering, $^2$INMC \& IPAI \\
Seoul National University, Republic of Korea\\
\email{\{jeeit17, seohoiki3215, sychun\}@snu.ac.kr}}
\maketitle
\def\thefootnote{*}\footnotetext{Authors contributed equally. $^{\dagger}$ Corresponding author.}

\appendix

\section{Additional Experimental Details\label{exep_detail}}

\subsection{Dataset}
\subsubsection{Dataset description.}
AGD20K~\cite{luo2022learning} is an affordance grounding dataset, consisting of 20,061 exocentric images and 3,775 egocentric images that are categorized based on object and interaction labels. During evaluation, egocentric images and interaction labels are provided to identify the most relevant regions of interaction in object images. Note that Ground Truth (GT) masks of AGD20K were annotated by the interactions between humans and objects in the OPRA dataset~\cite{fang2018demo2vec}. 

\subsubsection{Seen setting.}
The `Seen' setting of the AGD20K dataset comprises 36 \emph{interaction} labels and the train and test sets each contains 50 object categories.

\subsubsection{Unseen setting.}
The `Unseen' setting of the AGD20K dataset includes 25 \emph{affordance} categories. Unlike the `Seen' setting, the object classes in train and test sets do not overlap so that verification of whether the network can infer affordances for previously unseen objects is possible. There are 35 object classes in the train set and 12 object classes in the test set as the following object categories:
\begin{itemize}
    \item[$\bullet$] \textbf{Train set:} \textit{apple, badminton racket, baseball, baseball bat, basketball, bench, book, bottle, bowl, carrot, cell phone, chair, couch, discus, fork, frisbee, hammer, hot dog, javelin, keyboard, knife, microwave, motorcycle, orange, oven, punching bag, rugby ball, scissors, skateboard, snowboard, suitcase, surfboard, tennis racket, toothbrush, wine glass}
    \item[$\bullet$] \textbf{Test set:} \textit{axe, banana, bed, bicycle, broccoli, camera, cup, golf clubs, laptop, refrigerator, skis, soccer ball}
\end{itemize}

\subsection{Metrics}
\label{sec:metric}
Unlike segmentation tasks where GT is usually a binary mask, affordance grounding involves mapping the probability of an action on an object, thereby necessitating a probabilistic representation for GT. Following previous works~\cite{luo2022grounded, luo2022learning, luo2023learning, li2023locate, nagarajan2019grounded, chen2023affordance, yang2023grounding}, we employ Kullback-Leibler divergence (KLD)~\cite{KLD}, similarity (SIM)~\cite{SIM}, and normalized scanpath saliency (NSS)~\cite{NSS} as metrics to evaluate our method where KLD quantifies the discrepancy between two probability distributions, SIM measures the similarity between two distributions, and NSS evaluates the correspondence between two maps. The details for computing these metrics are as follows. 
Firstly, we resize the GT mask $\mathcal{M}_{GT}$ and the model's predicted attention map $\mathcal{M}_{aff}$ to $224\times224$ using bilinear interpolation so that $\{ \mathcal{M}_{GT}, \mathcal{M}_{aff} \} \in \mathbb{R}^{224\times224}$. Then, they are min-max normalized and each element is divided by the sum of all elements to obtain $\hat{\mathcal{M}}_{GT}$ and $\hat{\mathcal{M}}_{aff}$ as follows:
\begin{equation}
\centering
    \hat{\mathcal{M}}_{GT} = \mathcal{M}_{GT} / \sum{\mathcal{M}_{GT}}, \; \hat{\mathcal{M}}_{aff} = \mathcal{M}_{aff} / \sum{\mathcal{M}_{aff}}.
    \label{eq1:normalization}
\end{equation} 
Lastly, Using $\hat{\mathcal{M}}_{GT}$ and $\hat{\mathcal{M}}_{aff}$, KLD and SIM are calculated as following:
\begin{equation}
\centering
    \mathrm{KLD}(\hat{\mathcal{M}}_{GT} || \hat{\mathcal{M}}_{aff}) = \sum{\hat{\mathcal{M}}_{GT}} \cdot \log\left(\frac{\hat{\mathcal{M}}_{GT}}{\hat{\mathcal{M}}_{aff}}\right),
    \label{eq2:KLD}
\end{equation}
\begin{equation}
\centering
    \mathrm{SIM}(\hat{\mathcal{M}}_{GT}, \hat{\mathcal{M}}_{aff}) = \sum{\textnormal{min}(\hat{\mathcal{M}}_{GT}, \hat{\mathcal{M}}_{aff})}.
    \label{eq3:SIM}
\end{equation}
The following calculations are conducted to compute NSS:
\begin{equation}
\centering
    \Tilde{\mathcal{M}}_{GT} = \mathbf{1}(\mathcal{M}_{GT} > 0.1), \ \Tilde{\mathcal{M}}_{aff} = \frac{\mathcal{M}_{aff} - \mu_{\mathcal{M}_{aff}}}{\sigma_{\mathcal{M}_{aff}}}
    \label{eq4:NSS_pre}
\end{equation}
where $\mathbf{1}(\cdot)$ denotes the indicator function, $\mu_{\mathcal{M}_{{aff}}}$ and $\sigma_{\mathcal{M}_{{aff}}}$ represent the mean and standard deviation of $\mathcal{M}_{aff}$, respectively. From these, we calculate NSS as follows:
\begin{equation}
\centering
    \mathrm{NSS}(\Tilde{\mathcal{M}}_{GT}, \Tilde{\mathcal{M}}_{aff}) = \frac{1}{\sum{\Tilde{\mathcal{M}}_{GT}}}\sum{\Tilde{\mathcal{M}}_{GT} \cdot \Tilde{\mathcal{M}}_{aff}}.
    \label{eq5:NSS}
\end{equation}

\subsection{User study}
Affordances can be ambiguous due to several factors, including context dependence, perceptual limitations, and subjective interpretations. Therefore, we undertook a comprehensive assessment of our INTRA's prediction output through a user study. The study addressed three aspects: `validity', which evaluates fidelity of the affordance grounding map, `finesse', which evaluates granularity and detail of the presented affordance map, and `separability', which evaluates the model's ability to appropriately ground the same object in different locations for various interactions. Eight interactions (\emph{`push', `cut with', `take photo', `sip', `open', `sit on', `pour', `hold'}) and nine objects (\emph{`motorcycle', `scissors', `camera', `cup', `microwave', `refrigerator', `bicycle', `wine glass', `knife'}) were chosen at random. Respondents assigned scores ranging from 1 to 5 to the affordance maps generated by prior arts (Cross-View-AG+~\cite{luo2022grounded}, LOCATE~\cite{li2023locate}), GT itself, and our proposed INTRA (Ours). The sequence of results from each model was randomized for each questionnaire. A total of 104 respondents evaluated the 4 models (including the oracle or GT) across the 9 items.

\subsection{Experiment on additional datasets}
In addition to AGD20K, our INTRA was evaluated on three additional datasets: IIT-AFF~\cite{nguyen2017object}, CAD~\cite{sawatzky2017weakly}, and UMD~\cite{myers2015affordance}. While these datasets were originally created for affordance segmentation (not specifically for affordance grounding), they can still provide valuable evaluations and comparisons on how well our INTRA performs affordance grounding in terms of accuracy and finesse compared to other prior arts on generalized datasets. We made the binary segmentation mask using a threshold of 0, thus setting the mask value to be 1 if the value in the mask exceeds 0. Since prior arts in affordance grounding~\cite{li2023locate, luo2022grounded, luo2022learning} can not predict the affordances that were not part of the training data, the affordances that were not included in the train set were excluded from the dataset for fair comparisons (advantageous for prior works). With the modified dataset, we assessed the models with a total of 9,797 images from IIT-AFF, 20,016 images from CAD, and 39,846 images from UMD. Since the majority of objects presented in these datasets are also included in the AGD20K dataset, we conducted evaluations utilizing a model trained under the `Seen' setting.  All remaining evaluation processes were carried out in the same manner as described in Sec~\ref{sec:metric}.

\section{Additional Pipeline Details}

\subsection{Additional details on network architecture}
\subsubsection{Design choices for network architecture.}
$\mathcal{F}_{exo}$ was extracted using the pre-trained DINOv2-base~\cite{oquab2023dinov2} as an image encoder, with a patch size of 14 and a feature dimension of 768. For $\mathcal{F}_{text}$, we employed the text encoder of ALBEF~\cite{li2021align} based on BERT-base-uncased~\cite{devlin2018bert}. In the affordance map generation module, a transformer encoder architecture stacked with 4-layers and 4-head attention was utilized. Finally, the 2-layer convolution network in the module projects 768-channel feature into a single-channel $\mathcal{M}_{aff}$. For the projection layer, a 3-layer MLP with the output dimension of 128 was adopted to generate $z_{exo}$, which was later utilized in contrastive learning.
\begin{figure*}[t]
    \includegraphics[width=\linewidth]{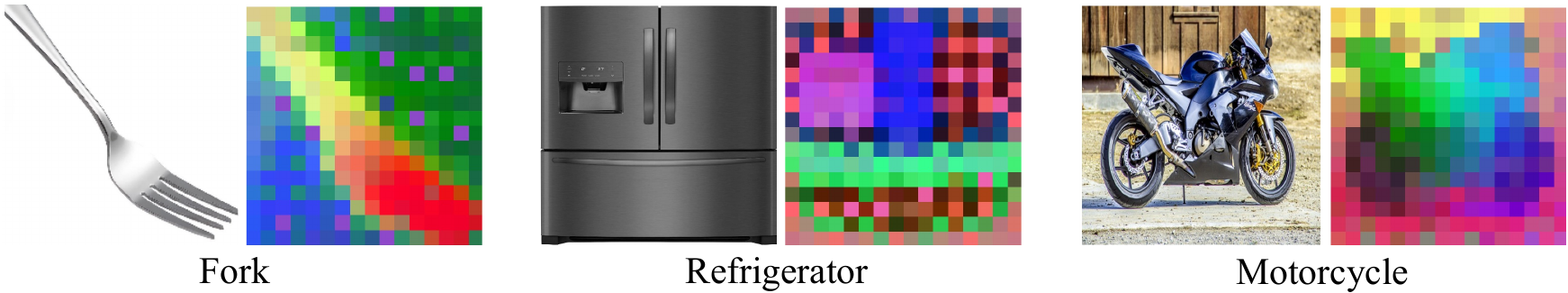}
    \vspace{-1em}
    \caption{PCA analysis of DINOv2~\cite{oquab2023dinov2} features on three objects: fork, refrigerator, motorcycle. As depicted in the figure, the DINOv2 is capable of distinguishing the tip and handle of the fork; the door of the refrigerator; the handle, saddle, and the wheel of the motorcycle.}
    \vspace{-1em}
    \label{fig:fig_PCA}
\end{figure*}
\subsubsection{Rationale for adopting DINO.}
DINOv2~\cite{oquab2023dinov2} was selected as the image encoder for our INTRA due to its superior performance in extracting low-level features from images. To visualize the ability of DINOv2, we projected DINOv2 image features onto 3 principal components using 3-dimensional PCA, and then normalized each dimension from 0 to 1. We then mapped the values to integers from 0 to 255 to obtain an RGB image. As depicted in Fig.~\ref{fig:fig_PCA}, DINOv2 is capable of distinguishing the object parts: the tip and handle of a fork, the door of a refrigerator, and the handle, saddle, and the wheel of a motorcycle.

\subsection{Additional details on usage of LLM}

\subsubsection{Text synonym augmentation.}
In text synonym augmentation, LLM was used to find synonyms. Examples of queries for finding synonyms can be found in Tab.~\ref{tab:query_text_synonym}. To provide the exact context of interaction, examples of interaction and object pairs were given. Three synonyms were found, and those that overlapped with labels of other classes were replaced before usage. The interaction names used in conditioning were replaced with generated synonyms with a 20\% probability, while the names of classes used in interaction relationship-guided contrastive learning remained unchanged. Examples of the synonyms used are depicted in Tab.~\ref{tab:example_text_aug}. ChatGPT-4~\cite{achiam2023gpt} was adopted as LLM, and no hyperparameters were used. 

\subsubsection{Query for interaction relationship map.}
We utilized LLM to determine whether the given affordance pair acts on the same part of an object, \textit{i.e.},  whether the pair is positive or not. LLM was provided with a few examples, comprising 1.4\% of the whole interaction pair, for the task and then asked to classify the given pair with Chain-of-Thoughts (CoT)~\cite{wei2022chain}.  The full query for the generation of interaction relationship map is provided in Tab.~\ref{tab:query}. ChatGPT-4~\cite{achiam2023gpt} was adopted as LLM, and no hyperparameter was used. 

\begin{table}[h!]
\caption{An example of LLM query and answer to find synonyms for text synonym augmentation.}
\begin{tabular}{ p{\dimexpr 0.1\linewidth}
                 p{\dimexpr 0.9\linewidth}}
\hline
\multicolumn{2}{l}{\textbf{Example of LLM query for text synonym augmentation}}                \\ \hline
User & Give me three synonyms for each verbs in form of [`word1', `word2', `word3']. \newline
1. `beat' in context of `beat drum'\newline
2. `boxing' in context of `boxing punching bag'\newline
3. `brush with' in context of `brush with toothbrush'\newline
4. `carry' in context of `carry suitcase'\newline
5. `catch' in context of `catch baseball'\newline
\\
LLM  & 1. [`strike', `hit', `pound'] \newline
2. [`punching', `sparring', `training'] \newline
3. [`clean with', `scrub with', `sweep with'] \newline
4. [`transport', `haul', `convey'] \newline
5. [`grab', `snag', `snatch']\newline
\\ \hline
\end{tabular}
\label{tab:query_text_synonym}
\end{table}

\begin{table}[!hp]
\caption{An example of synonyms generated by LLM for text synonym augmentation.}
\centering
\begin{adjustbox}{width=1\textwidth}
\begin{tabular}{ p{\dimexpr 0.15\linewidth}
                 p{\dimexpr 0.85\linewidth}}
\hline
\multicolumn{2}{l}{\textbf{Examples of synonyms for text synonym augmentation}}                                                                                      \\ \hline
beat  & strike, tap, pound \\
boxing & punching, sparring, kicking \\
brush with & clean with, scrub with, sweep with \\
carry & transport, haul, convey \\
catch & grab, snag, snatch \\
cut & slice, chop, carve \\
cut with & slice with, carve with, cut with \\
drag & pull, haul, tow \\
drink with & drink from, quaff from, imbibe from \\
eat & consume, devour, ingest \\
hit & strike, smack, blow \\
hold & grasp, embrace, grip \\
jump & skate, glide, roll \\
kick & kick out, boot, strike \\
lie on & rest on, recline on, lie upon \\
lift & hold, raise, hoist \\
look out & watch out, gaze out, observe \\
open & unzip, unpack, reveal \\
pack & bundle, assemble, prepare \\
peel & peel off, strip off, skin \\
pick up & pick up, collect, gather \\
pour & flow, spill, dispense \\
push & shove, press, thrust \\
ride & cycle, pedal, roll \\
sip & nibble, salute with, taste \\
sit on & sit upon, rest on, perch on \\
stick & pike with, thrust with, jab with \\
stir & mix, whisk, blend \\
swing & swipe, swat, clout \\
take photo & photograph with, capture with, shoot with \\
talk on & speak on, communicate on, converse on \\
text on & message on, compose on, write on \\
throw & toss, fling, hurl \\
type on & key in, enter with, type with \\
wash & rinse, scrub, cleanse \\
write & record with, scrible with, jot with \\
\hline
\end{tabular}
\end{adjustbox}
\label{tab:example_text_aug}
\end{table}
\noindent

\newpage
\begin{small}
\begin{longtable}[!hp]{p{\dimexpr 0.1\linewidth} p{\dimexpr 0.9\linewidth}}
\caption{An example of LLM query and answer to determine whether [`drink with', `sip'] pair is `positive' or `negative' pair.}
\label{tab:query}
\\ \hline
\multicolumn{2}{l}{\textbf{Example of LLM query for interaction-relationship map generation}} \\ \hline
User & You should clarify each verb's relation whether it is positive pair or negative pair. Here's some example of the examples of clarification. \newline
ex1) `hold' and `carry' is `positive' pair. Because if we do the given interaction to the object like suitcase, ski, or snowboard, we interacts with the same part of the object. The side part of the object for `hold' and `carry'. \newline
ex2) `hit' and `carry' is `positive’ pair. Because if we do the given interaction to the object like baseball bat or badminton racket, we interacts with the same part of the object. The thin part of the object for `hit' and `carry’. \newline
ex3) `cut with’ and `hold’ is `negative’ pair. Because if we `cut with’ the object like scissor, or knife, we interacts with the different part of the object. Sharp part of the object for `cut with’ and `hold’ for handle or dull part. \newline
ex4) `sip’ and `hold’ is `negative’ pair. Because if we do the given interaction to the object like cup, bottle, or wine glass, we interacts with the different part of the object. Rim of the object for `sip’, and `hold’ for handle or round part. \newline
\newline
Think of 5 objects that can be commonly interacted with. For each object, describe the interactions and then list the object parts that should be interacted with step by step. Use this information to create criteria for classification based on the given example.
Specify the final answer with short format and it should be one of [positive, negative]. [`drink with', `sip'] $\rightarrow$ [`positive', `negative']? \\ \\
LLM & 1. Cup
\begin{itemize}
    \item Drink with: When using a cup to drink, the action involves bringing the rim of the cup to your lips and tilting it to allow the liquid to flow into your mouth. 
    \item Sip: Sipping from a cup also specifically involves using the rim. The key distinction is in the amount of liquid and the care with which it's consumed, but it's the same part of the cup — the rim — that facilitates the action.
\end{itemize}
 2. Straw
\begin{itemize}
    \item Drink with: Drinking with a straw involves placing one end of the straw in your mouth and the other end in the beverage. The interaction is exclusively with the straw as the medium through which the beverage is drawn up.
    \item Sip: Sipping through a straw also involves the same exact use of the straw. The action might be more gentle or controlled, but it still requires using the straw to draw the liquid up in the same manner.
\end{itemize} \\
& 3. Glass
\begin{itemize}
    \item Drink with: Using a glass to drink involves directly interacting with the rim of the glass. You typically tilt the glass to pour the beverage into your mouth, engaging directly with the rim.
    \item Sip: Sipping from a glass, much like drinking, involves delicate interaction with the rim of the glass. The rim is the point of contact for controlling the intake of the liquid, even if the sipping is slower or in smaller quantities.
\end{itemize}
4. Bottle
\begin{itemize}
    \item Drink with: Drinking from a bottle involves placing your lips around the opening or the rim of the bottle. The beverage is consumed by tilting the bottle and allowing the liquid to flow into your mouth through the opening.
    \item Sip: Sipping from a bottle uses the same opening. The primary interaction is still with the rim or the mouth of the bottle, with the main difference being the amount of liquid consumed and the rate at which it is consumed.
\end{itemize}
5. Tea Cup
\begin{itemize}
    \item Drink with: Drinking from a tea cup involves using the rim of the tea cup, typically held by a handle for ease. The rim is the point of contact with the mouth, and the tea is consumed by tilting the cup.
    \item Sip: Sipping from a tea cup also involves the rim. While sipping might suggest a more refined or controlled action, especially with hot beverages like tea, it still requires interaction with the same part of the tea cup — the rim.
\end{itemize}
In each case, drink with and sip utilize the same part of these objects, confirming that they form a positive pair based on their interaction with the same part of the object. 
\\ \hline
\end{longtable}
\end{small}

\section{Additional Ablation Studies}
\subsection{Ablation study on interaction relationship map\label{interaction_realtionship_map}}
The ablation study in the main paper includes quantitative results with various relationship maps. $\mathcal{R}$ generated with LLM (Ours), was substituted with  $\mathcal{R}_{WordNet}$, $\mathcal{R}_{Co-Occur.}$, and $\mathcal{R}_{Word2Vec}$. $\mathcal{R}_{WordNet}$ was generated by calculating Wu-Palmer similarities~\cite{wu-palmer-1994-verb} for each interaction pair using WordNet~\cite{miller1995wordnet}. $\mathcal{R}_{Co-Occur.}$ was generated using the co-occurence probability of each interaction pair, which is the inner product of word vectors from the GloVe~\cite{pennington2014glove} 840B model. $\mathcal{R}_{Word2Vec}$ contains the similarity of each interaction pair calculated with Word2Vec~\cite{mikolov2013efficient}. All matrices were converted to binary with threshold of 0.5. 

\subsection{Ablation study on text encoder\label{text_encoder}}
A text encoder was employed to enhance the flexibility to input interactions and improve robustness against unseen interactions by leveraging VLM's text encoder, recognizing the intimate relationship between affordance grounding and visual information. We assessed the performance of various text embedding methods by integrating each encoder into our architecture, as detailed in Tab.~\ref{tab:text_encoder}. It is evident that while random embedding under nearly orthogonal conditions yields satisfactory performance,  it struggles to infer novel interactions. On the other hand, employing BERT~\cite{devlin2018bert} enables the inference of novel interactions, although the ALBEF~\cite{li2021align} text encoder demonstrates superior performance.
\begin{table}[h!]
\caption{Quantitative results of ablation study on various text encoders. For the `Random', a 768-dimensional random vector was initialized from a Gaussian distribution for each interaction. Just as in the INTRA model, only the class token of the BERT~\cite{devlin2018bert} embedding was employed for `BERT'. The results indicate that the ALBEF~\cite{li2021align} text encoder outperforms others, because it embeds visual information in the text.}
\centering
\begin{tabular}{ccccc}
\hline
                        & Encoder & mKLD$\downarrow$  & mSIM$\uparrow$  & mNSS$\uparrow$  \\ \hline
\multirow{3}{*}{\rotatebox[origin=c]{90}{Seen}}   & Random  & 1.230 & 0.392 & 1.209 \\
                        & BERT~\cite{devlin2018bert}    & 1.286 & 0.389 & 1.138 \\
                        & ALBEF~\cite{li2021align}   & 1.199 & 0.407 & 1.239 \\ \hline
\multirow{3}{*}{\rotatebox[origin=c]{90}{Unseen}} & Random  & 1.520 & 0.319 & 1.118 \\
                        & BERT~\cite{devlin2018bert}    & 1.458 & 0.321 & 1.190 \\
                        & ALBEF~\cite{li2021align}   & 1.365 & 0.375 & 1.209 \\ \hline
\end{tabular}
\label{tab:text_encoder}
\end{table}

\subsection{Ablation study on the number of projection layers}
We conducted ablation experiments on the number of projection layers. The quantitative results can be found in Tab.~\ref{tab:projection layer}. The importance of the projection layer in contrastive learning is already well-known in prior works~\cite{chen2020simple, xue2024investigating, chen2021exploring}. Based on these experimental results, we used three projection layers for our pipeline. 

\begin{table}[]
\caption{Quantitative results of ablation study on the number of projection layers. We changed the number of projection layers, trained INTRA and compared the performance. Based on this experimental results, we used three projection layers for our pipeline.}
\centering
\begin{tabular}{ccccc}
\hline
\multicolumn{1}{l}{}    & \multicolumn{1}{l}{\# of proj. layers} & mKLD$\downarrow$  & mSIM$\uparrow$  & mNSS$\uparrow$  \\ \hline
\multirow{4}{*}{\rotatebox[origin=c]{90}{Seen}}   & 1                                      & 1.380 & 0.393 & 1.085 \\
                        & 2                                      & 1.260 & 0.401 & 1.157 \\
                        & 3                                      & 1.199 & 0.407 & 1.239 \\
                        & 4                                      & 1.223 & 0.400 & 1.198 \\ \hline
\multirow{4}{*}{\rotatebox[origin=c]{90}{Unseen}} & 1                                      & 1.680 & 0.298 & 0.891 \\
                        & 2                                      & 1.593 & 0.320 & 0.931 \\
                        & 3                                      & 1.365 & 0.375 & 1.209 \\
                        & 4                                      & 1.497 & 0.324 & 1.101 \\ \hline
\end{tabular}
\label{tab:projection layer}
\end{table}

\subsection{Ablation study on text synonym augmentation}
The detailed quantitative results of ablation study on the text synonym augmentation are provided in Tab.~\ref{tab:randomsynwhole} and Tab.~\ref{tab:randomsynnovel}. As shown in the table, text synonym augmentation enhances overall performance as well as the performance on novel verbs. The subsets `hold', `swing' and `take photo' were selected based on their proportion in the whole dataset to demonstrate that text synonym augmentation improves performance on novel verbs regardless of their proportion. `hold', `swing' and `take photo' each accounts for 24.17\%, 3.82\% and 2.45\% of the train set. 

\begin{table}[h!]
\caption{Quantitative results of ablation study on the text synonym augmentation. Text synonym augmentation enhances overall performance, especially in `Unseen' setting. `w/o aug.' denotes that the inference was done without text synonym augmentation and `w/ aug.' denotes that the inference was done with text synonym augmentation.}
\centering
\begin{tabular}{ccccccc}
\hline
         & \multicolumn{3}{c}{Seen} & \multicolumn{3}{c}{Unseen} \\
         & mKLD$\downarrow$          & mSIM$\uparrow$          & mNSS$\uparrow$          & mKLD$\downarrow$          & mSIM$\uparrow$           & mNSS$\uparrow$           \\ \hline
w/o aug. & 1.288        & 0.386       & 1.151       & 1.563        & 0.322        & 1.058        \\
w/ aug.  & 1.199        & 0.407       & 1.239       & 1.365        & 0.375        & 1.209        \\ \hline
\end{tabular}
\label{tab:randomsynwhole}
\end{table}

\begin{table}[h!]
\caption{Quantitative results of ablation study on the text synonym augmentation. Text synonym augmentation improves performance in novel verb inference, irrespective of their proportion in the train set.}
\centering
\begin{tabular}{cccccccccc}
\hline
         & \multicolumn{3}{c}{`hold'} & \multicolumn{3}{c}{`swing'} & \multicolumn{3}{c}{`take photo'} \\
         & mKLD$\downarrow$          & mSIM$\uparrow$         & mNSS$\uparrow$         & mKLD$\downarrow$          & mSIM$\uparrow$         & mNSS$\uparrow$         & mKLD $\downarrow$           & mSIM$\uparrow$           & mNSS$\uparrow$           \\ \hline
w/o aug. & 1.533       & 0.317       & 0.825       & 1.828        & 0.212       & 0.940       & 0.922         & 0.472         & 1.070         \\
w/ aug.  & 1.344       & 0.356       & 1.304       & 1.789        & 0.216       & 1.031       & 0.648         & 0.555         & 1.320         \\ \hline
\end{tabular}
\label{tab:randomsynnovel}
\end{table}

\begin{table}[h!]
\caption{Quantitative results of ablation study on the temperature of interaction relationship-guided contrastive loss. In temperature-scaled contrastive loss, selecting the appropriate temperature is crucial for achieving optimal model performance. Utilizing this property of contrastive loss, we conducted experiments to find a suitable value of temperature and trained INTRA with $\tau =0.2$.}
\centering
\begin{tabular}{ccccc}
\hline
                        & Temp. ($\tau$) & mKLD$\downarrow$  & mSIM$\uparrow$  & mNSS$\uparrow$  \\ \hline
\multirow{4}{*}{\rotatebox[origin=c]{90}{Seen}}   & 0.07           & 1.283 & 0.396 & 1.143 \\
                        & 0.1            & 1.263 & 0.384 & 1.173 \\
                        & 0.2            & 1.199 & 0.407 & 1.239 \\
                        & 0.4            & 1.291 & 0.382 & 1.137 \\ \hline
\multirow{4}{*}{\rotatebox[origin=c]{90}{Unseen}} & 0.07           & 1.462 & 0.333 & 1.126 \\
                        & 0.1            & 1.527 & 0.336 & 1.074 \\
                        & 0.2            & 1.365 & 0.375 & 1.209 \\
                        & 0.4            & 1.550 & 0.324 & 1.025 \\ \hline
\end{tabular}
\label{tab:temperature}
\end{table}

\begin{table}
\centering
\caption{Quantitative results of ablation study on the ratio of object-variance mitigation loss. Although this coefficient, $\lambda_{obj}$, does not have significant effect on performance of our model, $\lambda_{obj}=4$ works best for our network, as shown in the table.}
\begin{tabular}{ccccc}
\hline
                        & $\lambda_{obj}$ & mKLD$\downarrow$  & mSIM$\uparrow$  & mNSS$\uparrow$  \\ \hline
\multirow{4}{*}{\rotatebox[origin=c]{90}{Seen}}   
                        & 1           & 1.288 & 0.384 & 1.160 \\
                        & 2           & 1.249 & 0.395 & 1.196 \\
                        & 4            & 1.199 & 0.407 & 1.239 \\
                        & 8            & 1.288 & 0.390 & 1.155 \\ \hline
\multirow{4}{*}{\rotatebox[origin=c]{90}{Unseen}} 
                        & 1           & 1.554 & 0.302 & 1.083 \\
                        & 2            & 1.629 & 0.305 & 0.980 \\
                        & 4            & 1.365 & 0.375 & 1.209\\
                        & 8            & 1.624 & 0.294 & 0.939 \\ \hline
\end{tabular}
\label{tab:lambdaobj}
\end{table}

\subsection{Ablation study on hyperparameter\label{hyperparameter}}
\subsubsection{Temperature of interaction relationship-guided contrastive loss.}
In temperature-scaled contrastive loss, temperature is one of the important hyperparameters that determines the gap between positive and confusing negative samples. It is also well-known that the performance of the model varies depending on this value~\cite{wang2021understanding}. Specifically, in the affordance grounding task, setting the gap between hard negatives and positives is crucial due to instances where different parts of an object need to be localized in the same input image. Through exhaustive experiments, we have found a suitable value of temperature, and some of the results are presented in Tab.~\ref{tab:temperature}.

\subsubsection{Ratio of object-variance mitigation loss.}
The object-variance mitigation loss aims to mitigate the dissimilarity among object classes within a single interaction class. For example, the `hold' interaction class contains 21 different objects, such as `axe', `badminton racket', `baseball bat', `book', `bottle', `bowl' and `frisbee', each exhibiting distinct visual characteristics. Through experiments, we have determined an appropriate ratio for the object-variance mitigation loss. As depicted in Tab.~\ref{tab:lambdaobj}, there was no significant difference in performance, but $\lambda_{obj}=4$ yields the best results for our network.

\begin{figure*}[t!]
    \centering
    \includegraphics[width=\linewidth]{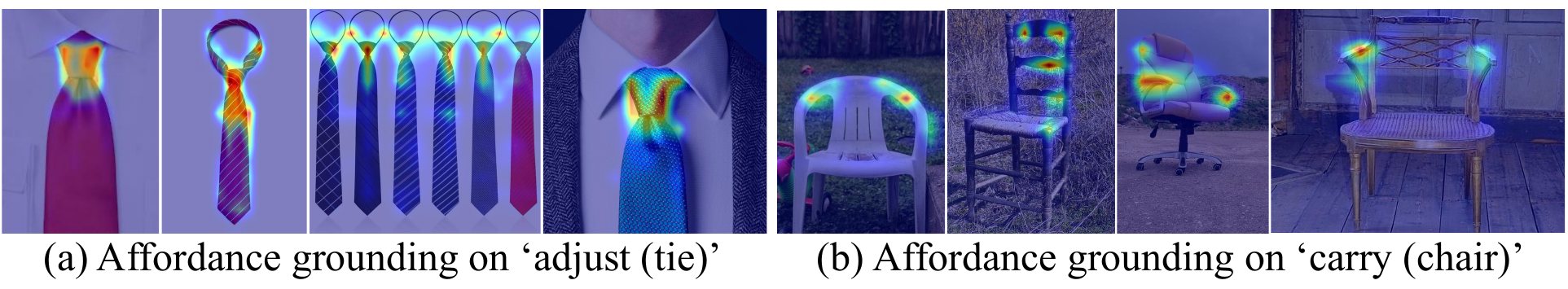}
    \caption{Qualitative results of INTRA trained with HICO-DET~\cite{chao:iccv2015}. (a) `adjust (tie)' accurately grounds the knot of the tie which human usually interacts with tie to `adjust' it. (b) `carry (chair)' precisely highlights the arm rest and back of the chair which are part of object facilitate for carrying.}
    \label{fig:fig_hico}
\end{figure*}

\section{Additional Experiments on Loss Design}
\subsubsection{Effectiveness of LLM usage.}
To generate an interaction relationship map, a comprehensive prior knowledge about interaction is required, as explained previously. While it is technically feasible to utilize a manually annotated matrix, it is suboptimal due to its scalability.  For a given number of interactions, denoted as $N_{\textnormal{inter}}$, the number of pairs that need to be determined grows quadratically as $_{N_{\textnormal{inter}}}C_2$, which follows a growth function of $O(N^2)$. For instance, while the AGD20K dataset features 36 interactions, necessitating determination of approximately 630 pairs, the HICO-DET~\cite{chao:iccv2015} dataset entails 6,786 pairs for 117 interactions, and 79,800 pairs for the SWiG-HOI~\cite{Pratt2020Swig, Wang_2021_ICCV} with 400 interactions. Therefore, the LLM-generated interaction relationship map is essential for the dataset scalability of the INTRA framework and the consistency of the pair classification. To demonstrate scalability of our method and the crucial role of LLM generated interaction-relationship map for dataset scalability, we trained our INTRA model with HICO-DET dataset. Since egocentric images or GT for quantitative evaluation are unavailable, we evaluated our trained network on crawled egocentric images containing interactions not present in AGD20K, but can be found in HICO-DET. As shown in Fig.~\ref{fig:fig_hico}, our network exhibits scalability for larger datasets and effectively identifies interactions such as `adjust', grounding the knot of a tie, and `carry', grounding the armrest and back of the chair. If our method was not based on LLM-generated interaction-relationship map, we had to manually annotate 6,786 interaction pairs to generate the map.

\subsubsection{Effectiveness of interaction relationship-guided contrastive loss.}
\begin{figure*}[t!]
    \centering
    \includegraphics[trim={0 0 15.5cm 0},clip,width=0.7\linewidth]{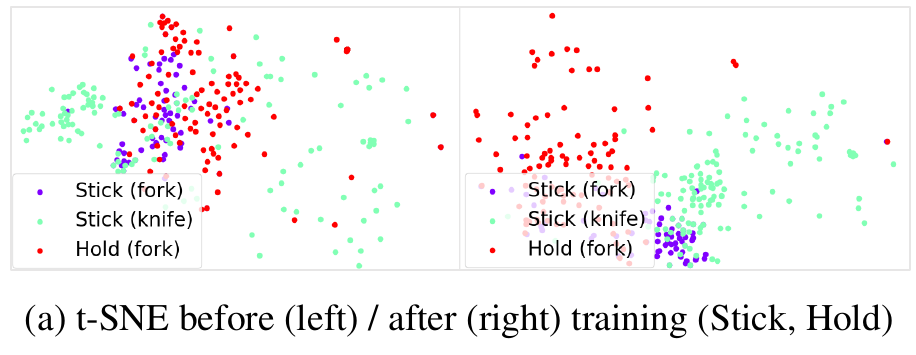}
    \includegraphics[trim={15.5cm 0 0 0},clip,width=0.7\linewidth]{figures_supp/supp_tsne.pdf}
    \caption{Feature distribution comparison before and after training was conducted using t-SNE. For t-SNE before training, the DINOv2 class token was utilized. For t-SNE after training, $f_{exo}$, which is simply the weighted sum of DINOv2 features, was used. As depicted in figure, features of interaction images implying different affordances(\textit{i.e.}, `stick' and `hold' fork or `push' and `sit on' bicycle) become separate as a result of interaction relationship-guided contrastive loss, while features of images containing similar object features still remain close to each other. }
    \label{fig:fig_tsne}
\end{figure*}

As seen in Fig.~\ref{fig:fig_tsne} (a) before training, DINOv2 features of `stick (fork)', `stick (knife)', and `hold (fork)' overlap, as `stick' usually comes with `hold'. However, since `stick' and `hold' fork imply different affordances, their features should be separated in the training. As training progresses, features of `stick' and `hold' forks being well separated, while `stick' forks and `stick' knife exhibit close feature distribution due to the closeness in their interactions, showing the effectiveness of interaction relationship-guided contrastive learning. Meanwhile, the object features contained in the images are well preserved, resulting in $f_{exo}$  of `stick' and `hold' fork still remain close to each other. The same tendency is also depicted in Fig.~\ref{fig:fig_tsne} (b), where `sit on' motorcycle becomes separate from `push' motorcycle after training. 
\section{Additional Results}
\subsection{Additional results on extended AGD20K testset}
In the main paper, we compared the performance using the existing AGD20K test set to maintain consistency with the experimental settings of previous works. However, we also compared the performance with LOCATE~\cite{li2023locate} using the recently extended AGD20K test set. Tab~\ref{tab:testsetextension} shows the quantitative comparison results of our method with LOCATE~\cite{li2023locate}. The results show that INTRA still outperformed in all metrics. 

\begin{table*}[t!]
\caption{Quantitative results on the new version of AGD20K for our method and LOCATE~\cite{li2023locate}. INTRA still outperformed LOCATE on all metrics.}
\centering
\begin{tabular}{ccccccc}
\hline
             & \multicolumn{3}{c}{Seen} & \multicolumn{3}{c}{Unseen} \\
             & mKLD$\downarrow$   & mSIM$\uparrow$   & mNSS$\uparrow$   & mKLD$\downarrow$    & mSIM$\uparrow$    & mNSS$\uparrow$   \\ \hline
LOCATE~\cite{li2023locate}       & 1.277  & 0.389  & 1.370  & 1.410        & 0.358        & 1.372       \\
\textbf{INTRA (Ours)} & \textbf{1.209}  & \textbf{0.443}  & \textbf{1.450}  & \textbf{1.388}        & \textbf{0.385}        & \textbf{1.384}       \\ \hline

\end{tabular}
\label{tab:testsetextension}
\end{table*}

\subsection{Affordance map visualization on exocentric images\label{exo_viz}}
\begin{figure*}[!p]
    \centering
    \includegraphics[width=\linewidth]{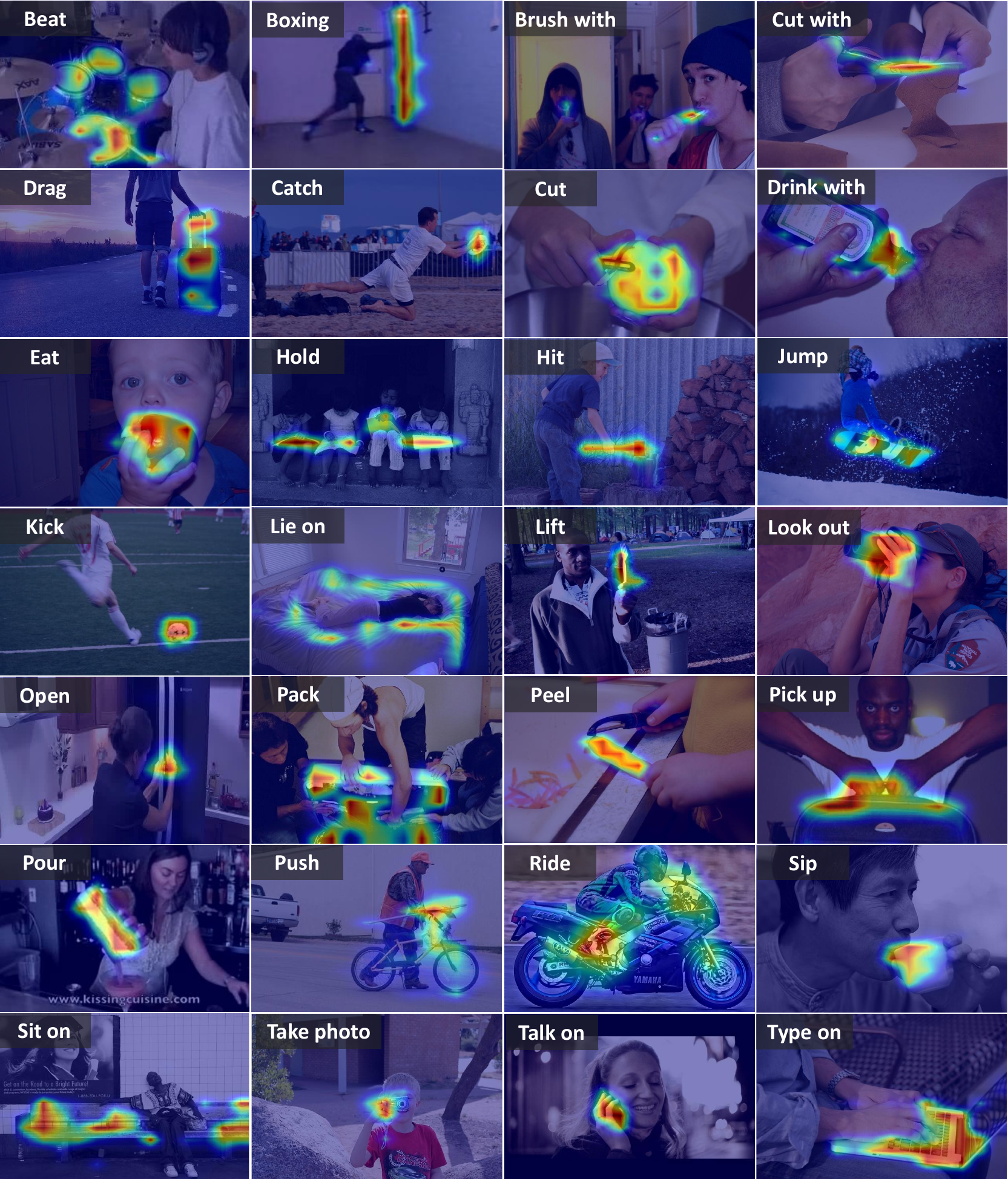}
    \caption{Affordance map visualization results on exocentric images. The model precisely localizes to the interacting part of the object rather than the person overall. Accurately pinpointing the part of the object involved in interaction is important, especially when occlusion occurs. Our model handles this well,  as shown in instances such as `hit', `look out', `open', `push', `ride', and `talk on'.}
    \label{fig:fig_exocentric}
\end{figure*}

 To verify that our proposed method, INTRA, focuses on the interaction-relevant object features in exocentric images during training, we conducted affordance map visualization on exocentric images. As illustrated in Fig.~\ref{fig:fig_exocentric}, even in complex scenes involving people interacting, it is evident that only the portion of the object engaged in interaction with people is grounded.

\subsection{Visualization of affordance grounding on the same objects with different interactions \label{sameobj_diffint}}
\begin{figure*}[!p]
    \centering
    \includegraphics[width=\linewidth]{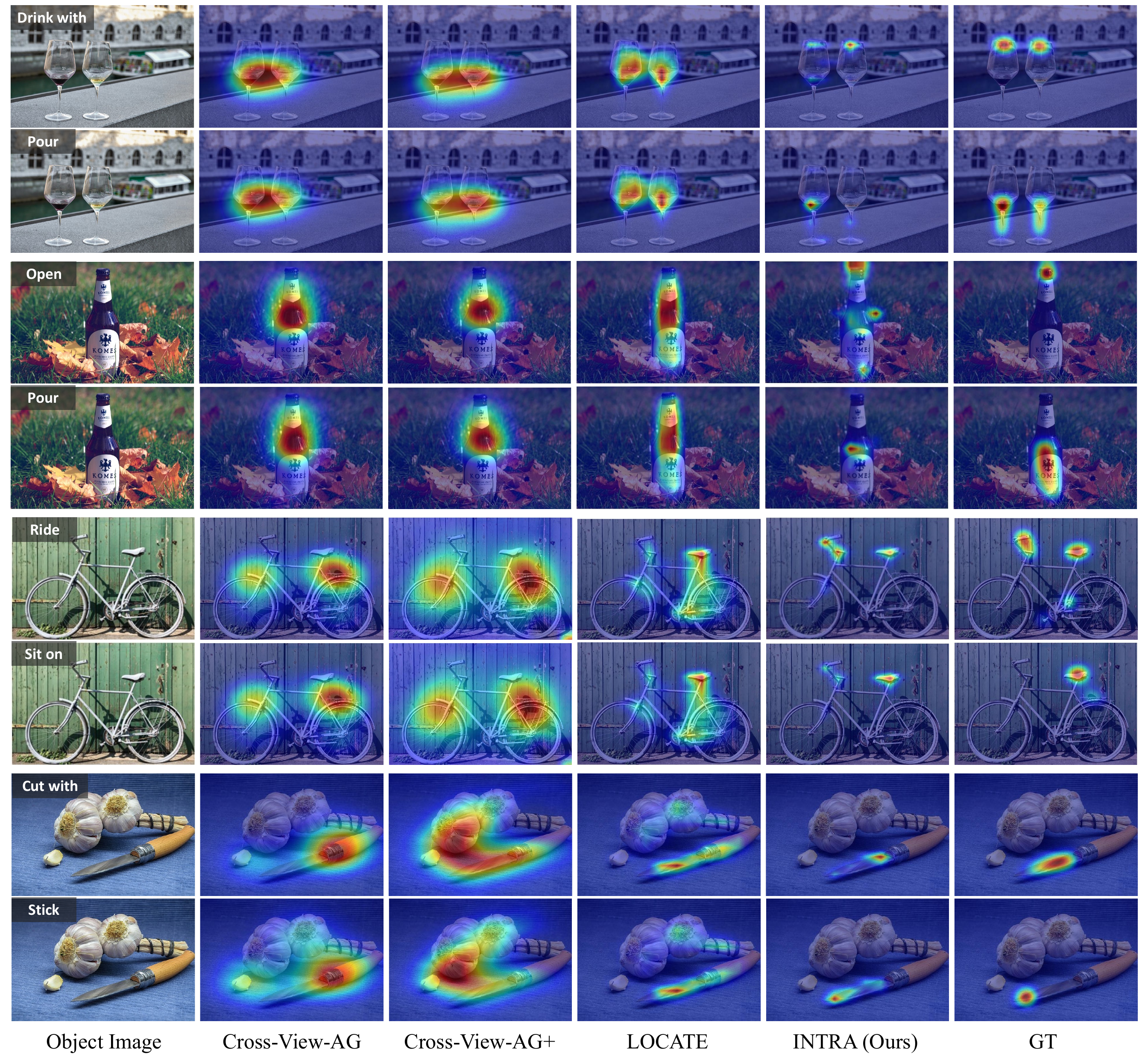}
    \caption{Visualization of affordance grounding on the same objects with different interactions compared to results of previous arts~\cite{luo2022grounded, luo2022learning, li2023locate}. We performed affordance grounding for `wine glass', `bottle', `bicycle', and `knife', where two interactions require different parts of objects to be grounded. The model grounded the rim of 'wine glass' for `drink with' and the handle for `pour'. For `bottle' and `bicycle', INTRA (Ours) precisely identifies the object parts corresponding to the given interactions. Particularly, for `knife', we observe that INTRA (Ours) grounds the blade for `cut with' and the tip of the knife for `stick'.}
    \label{fig:fig_sep}
\end{figure*}

Just as humans can interact with different parts of same objects for distinct interactions, it's crucial to accurately discern affordance grounding based on these interactions for further applications in real situations. We achieve this through interaction-guided contrastive learning and present various results in this section. As shown in Fig.~\ref{fig:fig_sep}, previous approaches struggle to accurately identify the object part that corresponds to the interaction, wheares INTRA (Ours) successfully grounds the relevant object part. We conducted the affordance grounding predictions for pairs such as `drink with' and `pour', `open' and `pour', `ride' and `sit on', and `cut with' and `stick', comparing the performance of previous methods with INTRA (Ours).

\subsection{Additional qualitative results on AGD20K dataset\label{additional_quali_agd20k}}
\begin{figure*}[!p]
    \centering
    \includegraphics[width=\linewidth]{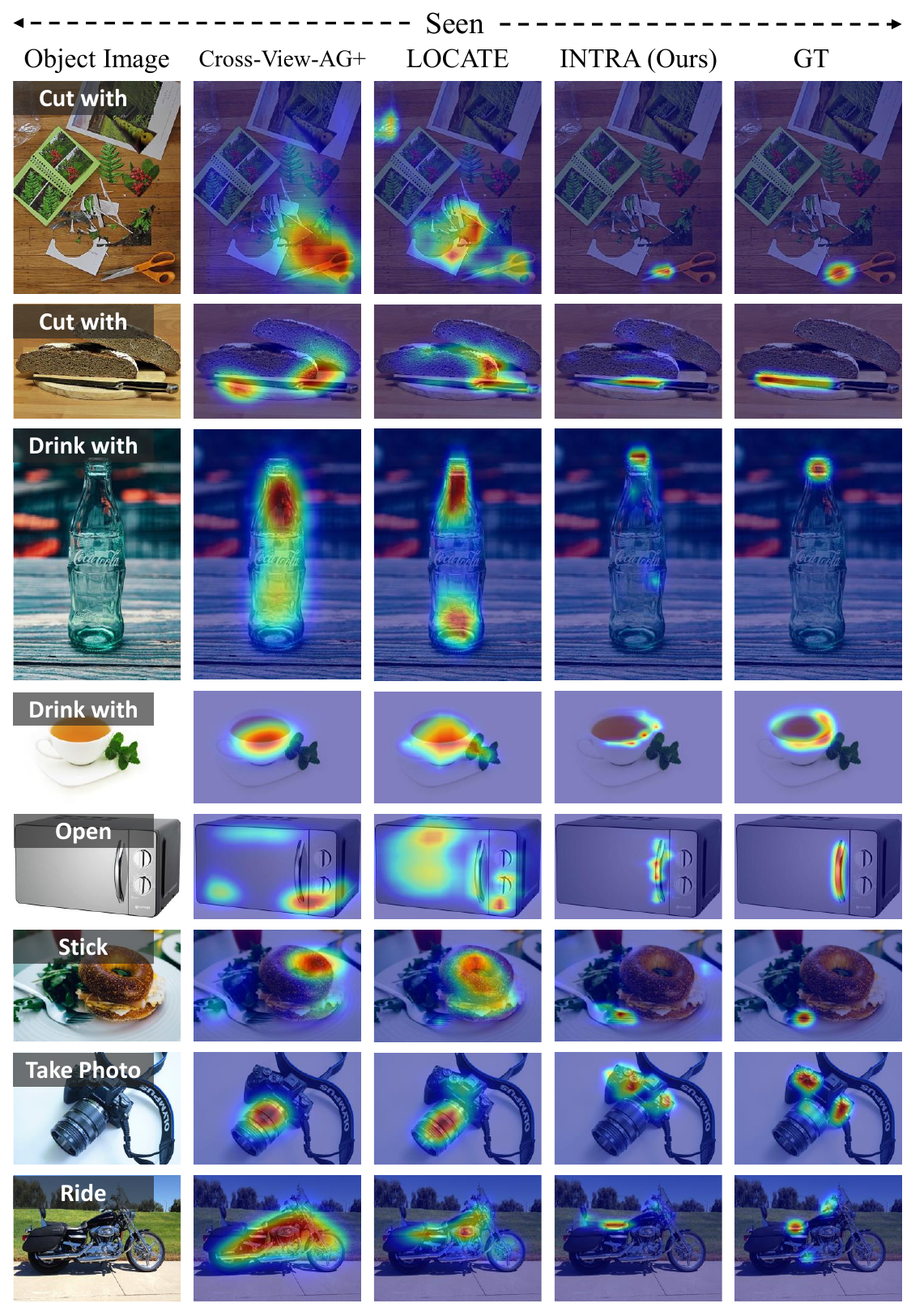}
    \caption{Additional qualitative results comparison between INTRA (Ours) and other baselines~\cite{luo2022grounded, luo2022learning, li2023locate} on `Seen' testset of AGD20K. INTRA (Ours) grounds affordance accurately when the images are clustered with many objects. For example, grounding results of `cut with' and `stick' accurately mark the blade of scissors, knife and the tip of fork. Unlike other baselines that tend to generate coarse heatmaps, our heatmaps are fine and localize only the relevant parts of the interactions.}
    \label{fig:fig_seen1}
\end{figure*}

\begin{figure*}[!p]
    \centering
    \includegraphics[width=\linewidth]{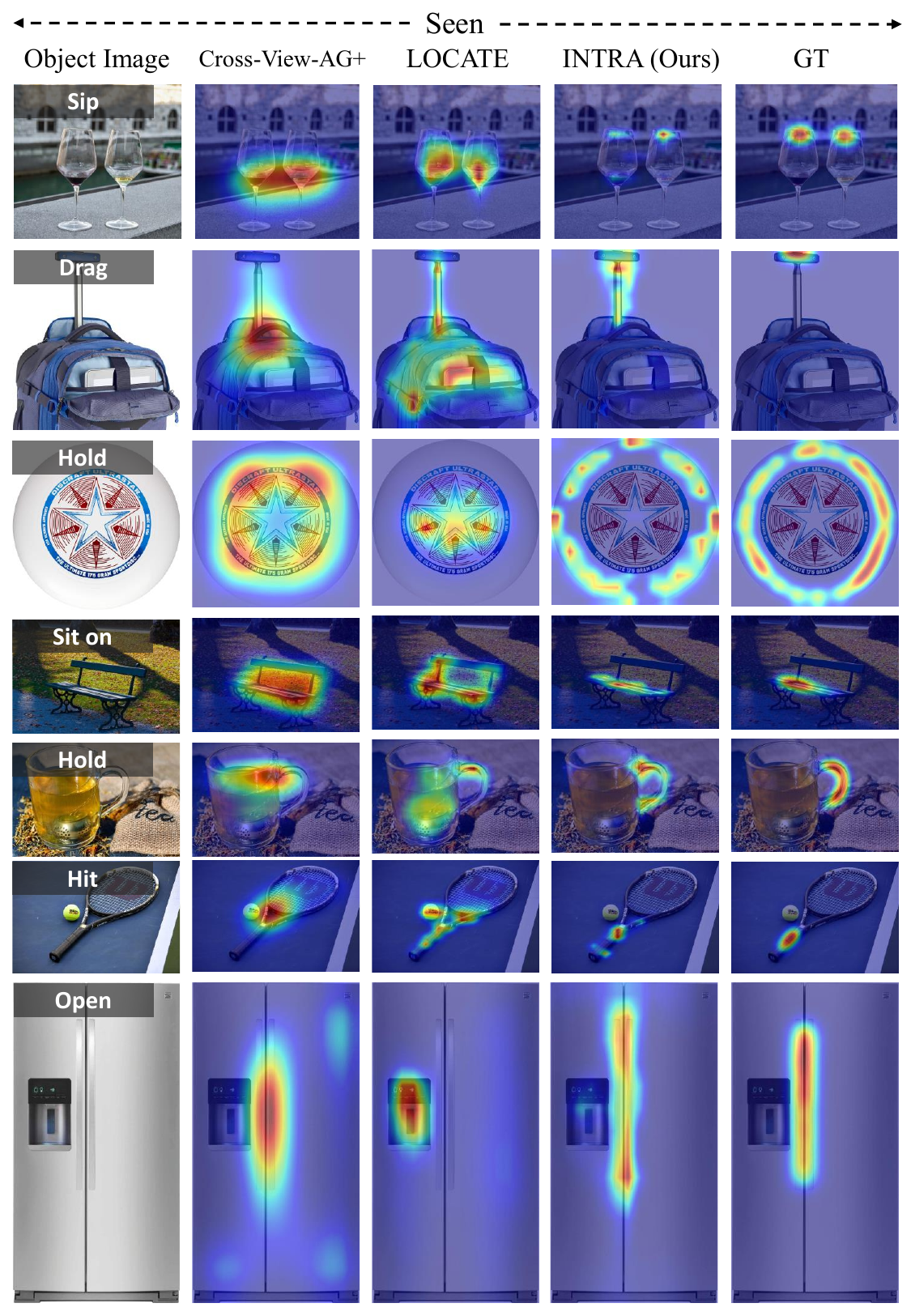}
    \caption{Additional qualitative results comparison between INTRA (Ours) and other baselines~\cite{luo2022grounded, luo2022learning, li2023locate} on `Seen' testset of AGD20K. INTRA (Ours) grounds affordance accurately and generates finer heatmaps. For example, grounding results of `drag' and `open' accurately mark the handle of suitcase, and the door handle of the refrigerator.}
    \label{fig:fig_seen2}
\end{figure*}

\begin{figure*}[!p]
    \centering
    \includegraphics[width=\linewidth]{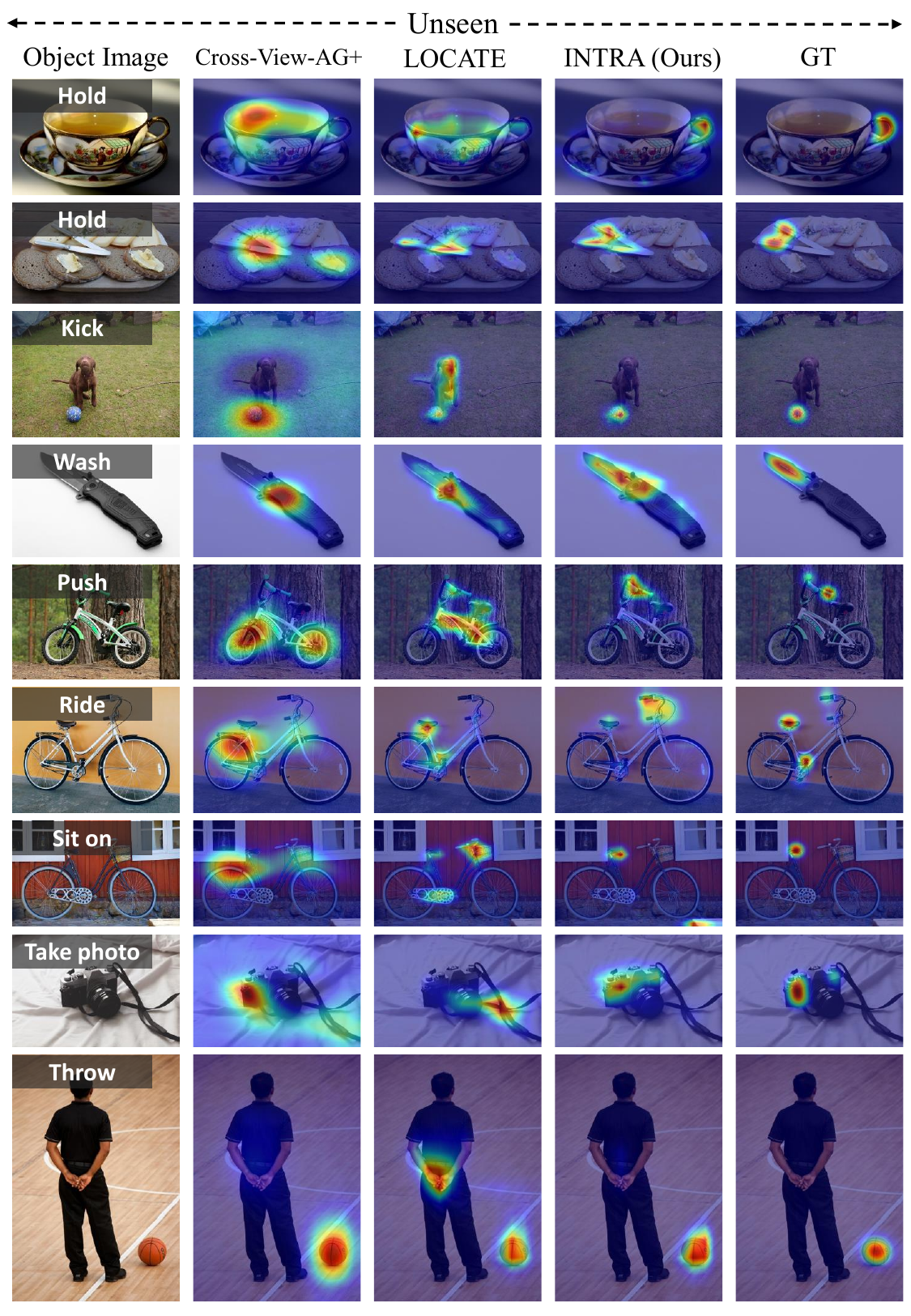}
    \caption{Additional qualitative results comparison between INTRA (Ours) and other baselines~\cite{luo2022grounded, luo2022learning, li2023locate} on `Unseen' testset of AGD20K. Qualitative comparison of grounding results in `Unseen' setting also shows that accuracy of our grounding results outperforms others. Especially, grounding result of `sit on' not only localizes bicycle saddle, but also the wooden chair that is partially shown in the image.}
    \label{fig:fig_unseen1}
\end{figure*}

\begin{figure*}[!p]
    \centering
    \includegraphics[width=\linewidth]{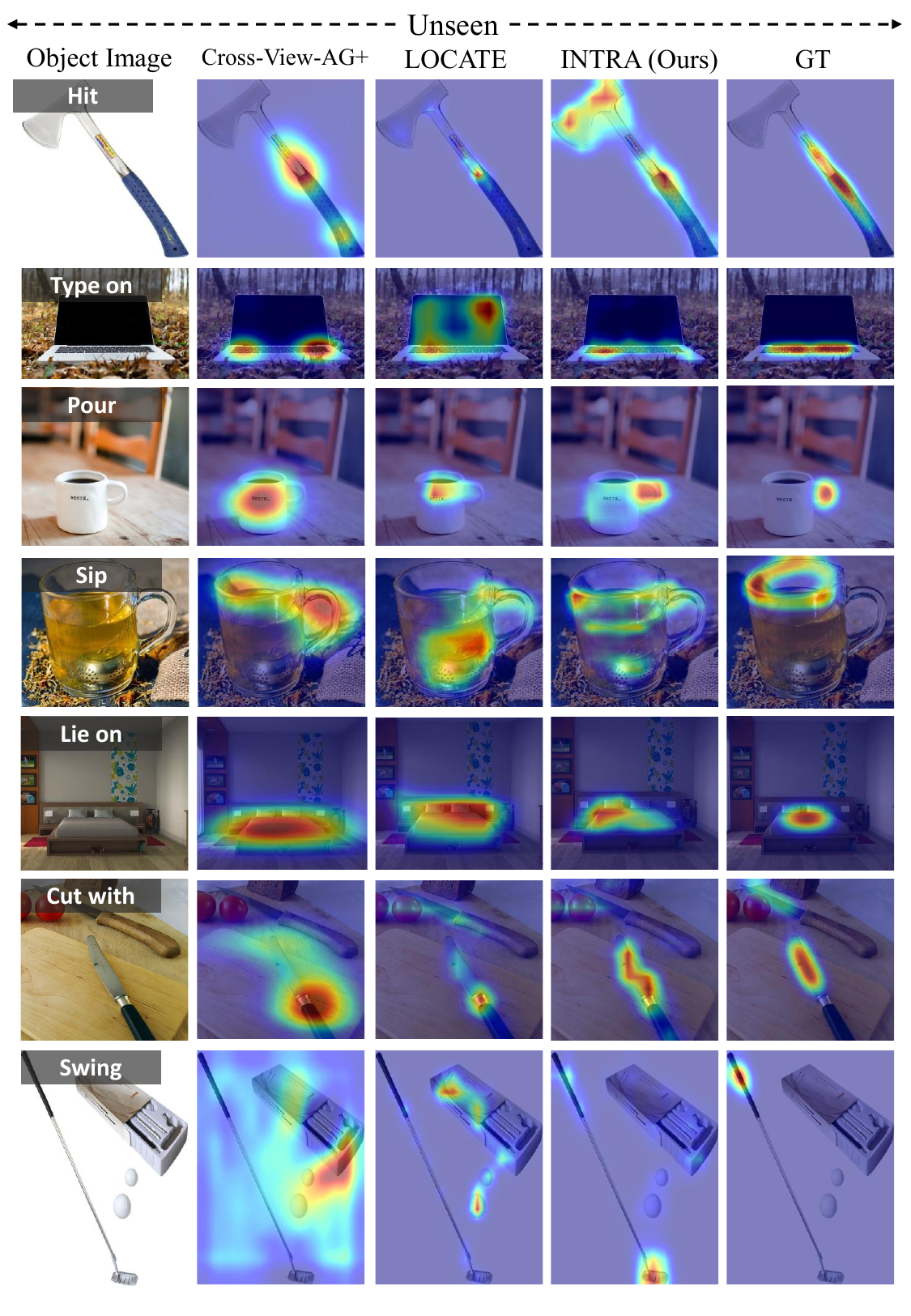}
    \caption{Additional qualitative results comparsion between INTRA (Ours) and other baselines~\cite{luo2022grounded, luo2022learning, li2023locate} on `Unseen' testset of AGD20K. Qualitative comparison of grounding results in `Unseen' setting also shows that accuracy of our grounding results outperforms others. Especially, the grounding result of `hit' not only localizes the handle of the axe, but also includes the blade of the axe, which can be interpreted as an integral part of the object incorporated with `hit'.}
    \label{fig:fig_unseen2}
\end{figure*}

In Fig.~\ref{fig:fig_seen1}, Fig.~\ref{fig:fig_seen2}, Fig.~\ref{fig:fig_unseen1} and Fig.~\ref{fig:fig_unseen2}, we present additional qualitative results and compare them with state-of-the art methods \cite{luo2022grounded, luo2022learning,li2023locate}. In both the `Seen' and `Unseen' setting, baselines struggles to accurately ground affordance when multiple objects are present in the images. Additionally, their heatmaps are also coarse and do not seem to handle occlusion that occurs in exocentric images. In contrast, even in complex scenes, INTRA (Ours) successfully grounds affordances. 

\subsection{Visualization of feasibility study on generalization property of INTRA}
\subsubsection{Domain gap.\label{dg}}
\begin{figure*}[!p]
    \centering
    \includegraphics[width=\linewidth]{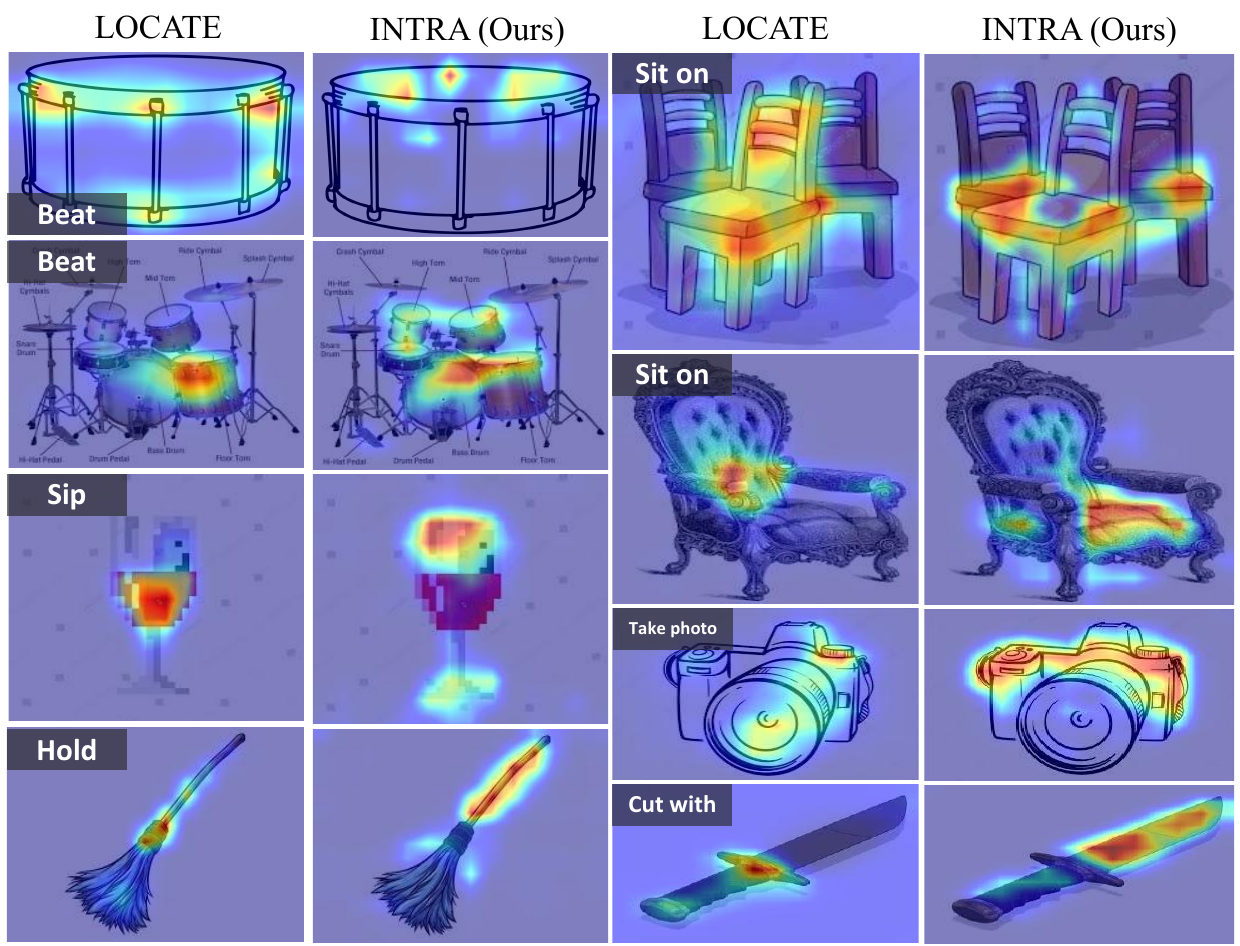}
\caption{Additional qualitative results comparison between INTRA (Ours) and other baseline\cite{li2023locate} on affordance grounding in object images with significant domain gap. INTRA (Ours) accurately and finely grounds affordances in pen-illustrated chairs, cameras and pixel-art wine glasses while other baseline can't. Also in case of `beat (drum)', while baseline model is inaccurately grounding the side of the drum, our model grounds center, top side of drum accurately. In case of drum set, our model grounded all drums that we can `beat' while the baseline grounded on the side of base drum.
The examples of `hold (broomstick)' and `cut with (knife)' shows that although there were significant domain gap between training images, INTRA (Ours) grounds the parts that are incorporated by interactions accurately.
}
    \label{fig:fig_dg}
\end{figure*}

Our framework, INTRA, demonstrates excellent grounding results when there is a significant domain gap between training and inference. We conducted inference using egocentric images from various domains to assess the robustness against this gap. We obtained synthetic images of objects present in the train set from the internet, including pen illustrations, pixel art, or object images from instructions.  Compared to other baseline model\cite{li2023locate}, our model accurately grounds affordances in those images. Especially, we can see in a picture of drum and chair of Fig.~\ref{fig:fig_dg} that our model successfully identifies all interaction-relevant regions of objects. Moreover, in images of pen-illustrated chair and camera, our model generates finer heatmaps. 

\begin{table*}[t!]
\caption{Comparison to previous arts on different object scales. Results of * are taken from ~\cite{li2023locate}. The test set is divided to `Big', `Medium' and `Small' based on the ratio of the mask to the whole image. INTRA (Ours) outperforms other baselines~\cite{mai2020erasing, pan2021unveiling, gao2021ts, nagarajan2019grounded, luo2022grounded, luo2022learning, li2023locate}  significantly in the `Small' subsets, while demonstrating competitive or superior performance in the `Medium' and `Big' subsets compared to the baselines.
}
\centering
\begin{tabular}{lcccccccccc}
\hline
\multicolumn{2}{c}{\multirow{2}{*}{Method}}                  & \multicolumn{3}{c}{Big}                                                                   & \multicolumn{3}{c}{Medium}                                                                & \multicolumn{3}{c}{Small}                                                                 \\ \cline{3-11}
\multicolumn{2}{c}{}                                         & \multicolumn{1}{c}{KLD$\downarrow$}            & \multicolumn{1}{c}{SIM$\uparrow$}            & NSS$\uparrow$            & \multicolumn{1}{c}{KLD$\downarrow$}            & \multicolumn{1}{c}{SIM$\uparrow$}            & NSS$\uparrow$            & \multicolumn{1}{c}{KLD$\downarrow$}            & \multicolumn{1}{c}{SIM$\uparrow$}            & NSS$\uparrow$             \\ \hline
\multicolumn{1}{l}{\multirow{8}{*}{{\rotatebox[origin=c]{90}{Seen}}}}   & EIL* \cite{mai2020erasing}            & \multicolumn{1}{c}{1.047}          & \multicolumn{1}{c}{0.461}          & 0.389          & \multicolumn{1}{c}{1.794}          & \multicolumn{1}{c}{0.284}          & 0.710          & \multicolumn{1}{c}{3.057}          & \multicolumn{1}{c}{0.123}          & 0.231          \\ 
\multicolumn{1}{l}{}                        & SPA* \cite{pan2021unveiling}           & \multicolumn{1}{c}{5.745}          & \multicolumn{1}{c}{0.317}          & 0.222          & \multicolumn{1}{c}{4.990}          & \multicolumn{1}{c}{0.228}          & 0.440          & \multicolumn{1}{c}{6.076}          & \multicolumn{1}{c}{0.118}          & 0.297          \\ 
\multicolumn{1}{l}{}                        & TS-CAM* \cite{gao2021ts}        & \multicolumn{1}{c}{1.039}          & \multicolumn{1}{c}{0.424}          & 0.166          & \multicolumn{1}{c}{1.814}          & \multicolumn{1}{c}{0.248}          & 0.401          & \multicolumn{1}{c}{2.652}          & \multicolumn{1}{c}{0.132}          & 0.352          \\ 

\multicolumn{1}{l}{}                        & Hotspots* \cite{nagarajan2019grounded}       & \multicolumn{1}{c}{0.986}          & \multicolumn{1}{c}{0.448}          & 0.408          & \multicolumn{1}{c}{1.738}          & \multicolumn{1}{c}{0.265}          & 0.672          & \multicolumn{1}{c}{2.587}          & \multicolumn{1}{c}{0.149}          & 0.683          \\  
\multicolumn{1}{l}{}                        & Cross-view-AG* \cite{luo2022learning}  & \multicolumn{1}{c}{\cellcolor{tabthird}0.766}         & \multicolumn{1}{c}{\cellcolor{tabthird}0.533}          & 0.652          & \multicolumn{1}{c}{1.485}          & \multicolumn{1}{c}{\cellcolor{tabthird}0.322}          & 1.040          & \multicolumn{1}{c}{\cellcolor{tabthird}2.373}          & \multicolumn{1}{c}{\cellcolor{tabthird}0.175}          & 0.927          \\  
\multicolumn{1}{l}{}                        & Cross-view-AG+* \cite{luo2022grounded} & \multicolumn{1}{c}{0.787}          & \multicolumn{1}{c}{0.521}          & \cellcolor{tabthird}0.660          & \multicolumn{1}{c}{\cellcolor{tabthird}1.481}          & \multicolumn{1}{c}{0.314}          & \cellcolor{tabthird}1.089          & \multicolumn{1}{c}{2.381}          & \multicolumn{1}{c}{0.167}          & \cellcolor{tabthird}0.959          \\ 
\multicolumn{1}{l}{}                        & LOCATE* \cite{li2023locate}         & \multicolumn{1}{c}{\cellcolor{tabfirst}0.676} & \multicolumn{1}{c}{\cellcolor{tabfirst}0.580} & \cellcolor{tabsecond}0.706          & \multicolumn{1}{c}{\cellcolor{tabfirst}1.178} & \multicolumn{1}{c}{\cellcolor{tabsecond}0.390}          & \cellcolor{tabfirst}1.316 & \multicolumn{1}{c}{\cellcolor{tabsecond}2.029}    & \multicolumn{1}{c}{\cellcolor{tabsecond}0.216}    & {\cellcolor{tabsecond}1.349}    \\  
\multicolumn{1}{l}{}                        & INTRA (Ours)    & \multicolumn{1}{c}{\cellcolor{tabsecond}0.695}    & \multicolumn{1}{c}{\cellcolor{tabsecond}0.579}    & \cellcolor{tabfirst}0.782 & \multicolumn{1}{c}{\cellcolor{tabsecond}1.193}    & \multicolumn{1}{c}{\cellcolor{tabfirst}0.394} & {\cellcolor{tabsecond}1.300}    & \multicolumn{1}{c}{\cellcolor{tabfirst}1.826} & \multicolumn{1}{c}{\cellcolor{tabfirst}0.239} & \cellcolor{tabfirst}1.587 \\ \hline
\multicolumn{1}{l}{\multirow{8}{*}{{\rotatebox[origin=c]{90}{Unseen}}}} & EIL* \cite{mai2020erasing} & \multicolumn{1}{c}{1.199}          & \multicolumn{1}{c}{0.393}          & 0.271          & \multicolumn{1}{c}{1.906}          & \multicolumn{1}{c}{0.246}          & 0.482          & \multicolumn{1}{c}{3.082}          & \multicolumn{1}{c}{0.113}          & 0.116          \\ 
\multicolumn{1}{l}{}                        & SPA* \cite{pan2021unveiling}            & \multicolumn{1}{c}{8.299}          & \multicolumn{1}{c}{0.259}          & 0.254          & \multicolumn{1}{c}{6.938}          & \multicolumn{1}{c}{0.186}          & 0.333          & \multicolumn{1}{c}{7.784}          & \multicolumn{1}{c}{0.095}          & 0.144          \\ 
\multicolumn{1}{l}{}                        & TS-CAM* \cite{gao2021ts}        & \multicolumn{1}{c}{1.238}          & \multicolumn{1}{c}{0.351}          & 0.072          & \multicolumn{1}{c}{1.970}          & \multicolumn{1}{c}{0.208}          & 0.236          & \multicolumn{1}{c}{2.766}          & \multicolumn{1}{c}{0.113}          & 0.124          \\ 
\multicolumn{1}{l}{}                        & Hotspots* \cite{nagarajan2019grounded}       & \multicolumn{1}{c}{1.015}          & \multicolumn{1}{c}{0.425}          & 0.548          & \multicolumn{1}{c}{1.872}          & \multicolumn{1}{c}{0.242}          & 0.605          & \multicolumn{1}{c}{2.693}          & \multicolumn{1}{c}{0.134}          & 0.544          \\ 
\multicolumn{1}{l}{}                        & Cross-view-AG* \cite{luo2022learning} & \multicolumn{1}{c}{0.884}          & \multicolumn{1}{c}{\cellcolor{tabthird}0.500}          & 0.728          & \multicolumn{1}{c}{\cellcolor{tabthird}1.595}          & \multicolumn{1}{c}{\cellcolor{tabthird}0.303}          & 0.945          & \multicolumn{1}{c}{\cellcolor{tabthird}2.558}          & \multicolumn{1}{c}{\cellcolor{tabthird}0.147}          & 0.692          \\ 
\multicolumn{1}{l}{}                        & Cross-view-AG+* \cite{luo2022grounded} & \multicolumn{1}{c}{\cellcolor{tabthird}0.867}          & \multicolumn{1}{c}{0.485}          & \cellcolor{tabthird}0.776          & \multicolumn{1}{c}{1.658}          & \multicolumn{1}{c}{0.279}          & \cellcolor{tabthird}0.988          & \multicolumn{1}{c}{2.630}          & \multicolumn{1}{c}{0.133}          & \cellcolor{tabthird}0.754          \\ 
\multicolumn{1}{l}{}                        & LOCATE* \cite{li2023locate}        & \multicolumn{1}{c}{\cellcolor{tabfirst}0.571}          & \multicolumn{1}{c}{\cellcolor{tabfirst}0.629}          & \cellcolor{tabfirst}0.956          & \multicolumn{1}{c}{\cellcolor{tabsecond}1.302}          & \multicolumn{1}{c}{\cellcolor{tabsecond}0.373}          & \cellcolor{tabfirst}1.257          & \multicolumn{1}{c}{\cellcolor{tabsecond}2.223}          & \multicolumn{1}{c}{\cellcolor{tabsecond}0.189}          & \cellcolor{tabsecond}1.071          \\ 
\multicolumn{1}{l}{}                        & INTRA (Ours)    & \multicolumn{1}{c}{\cellcolor{tabsecond}0.662}               & \multicolumn{1}{c}{\cellcolor{tabsecond}0.573}     &\cellcolor{tabsecond}0.955                & \multicolumn{1}{c}{\cellcolor{tabfirst}1.288}               & \multicolumn{1}{c}{\cellcolor{tabfirst}0.378}               & {\cellcolor{tabsecond}1.249}               & \multicolumn{1}{c}{\cellcolor{tabfirst}2.032}               &        \multicolumn{1}{c}{\cellcolor{tabfirst}0.230}   &\cellcolor{tabfirst}1.299     \\ \hline
\end{tabular}
\label{tab:diffscale}
\end{table*}

\subsubsection{Novel object.\label{no}}
\begin{figure*}[!p]
    \centering
    \includegraphics[width=\linewidth]{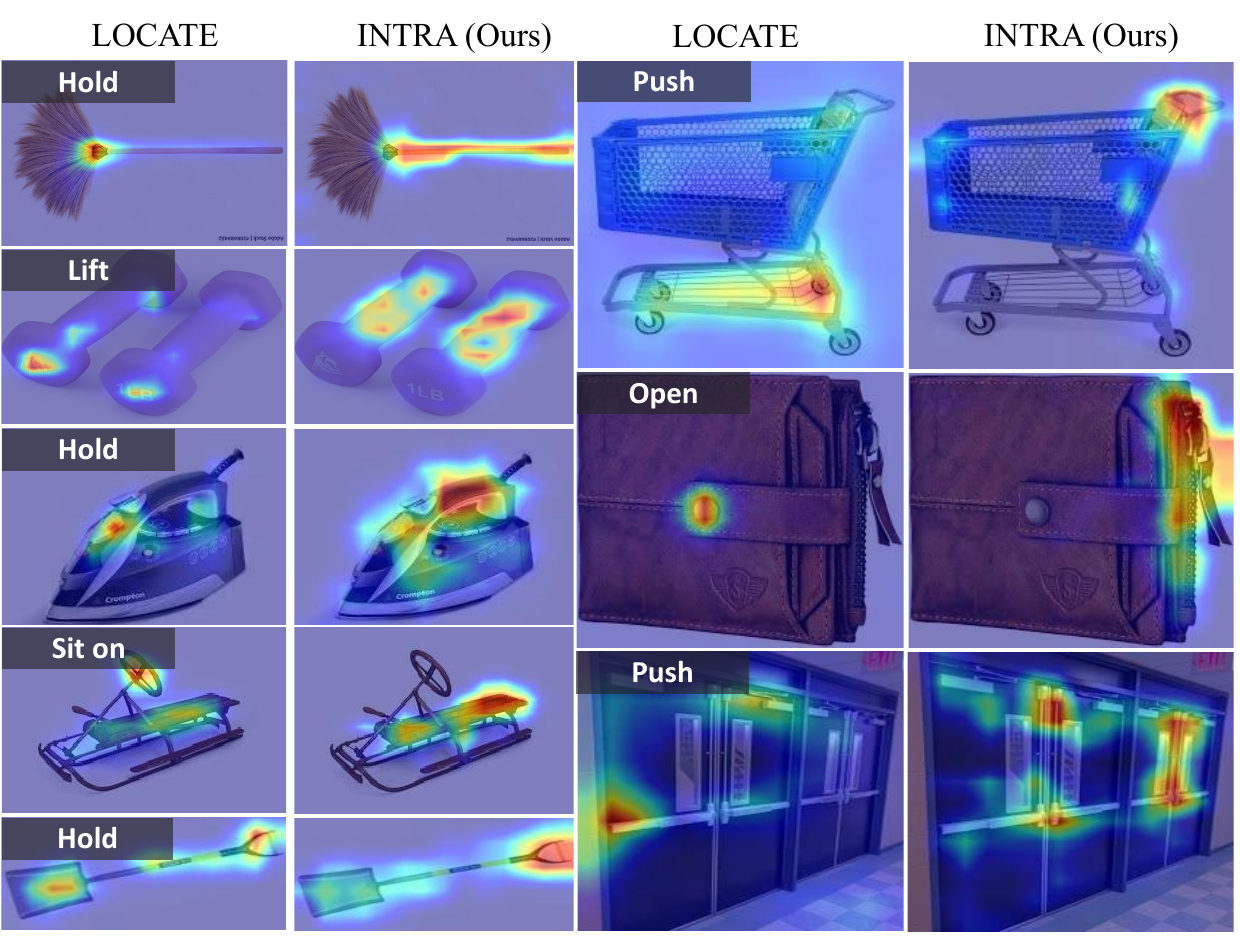}
\caption{Additional qualitative results comparison between INTRA (Ours) and other baseline\cite{li2023locate} on affordance grounding in novel objects. Our INTRA, as seen in examples like `open (wallet)' or `push (door)', accurately grounds more important interaction points such as center of doors or zipper of the wallet. Also, for example of `hold (shovel)', `hold (iron)' and `push (shopping cart)', it accurately captures the exact interaction points which are handle of the object. Although the train set does not contain images of weights or iron, INTRA (Ours) successfully grounds the interaction points.}
    \label{fig:fig_us_obj}
\end{figure*}

To analyze the generalization property of the affordance grounding, we conducted experiments on novel objects, which were not in the train set. We tested our model and baseline\cite{li2023locate} on novel objects that share common properties with objects in the train set, but have never been seen before. INTRA (Ours) successfully grounds the most interaction-relevant object parts, as demonstrated in examples of `wallet' and `door' in Fig.~\ref{fig:fig_us_obj}. 

\subsubsection{Novel interaction.\label{ni}}
\begin{figure*}[!p]
    \centering
    \includegraphics[width=\linewidth]{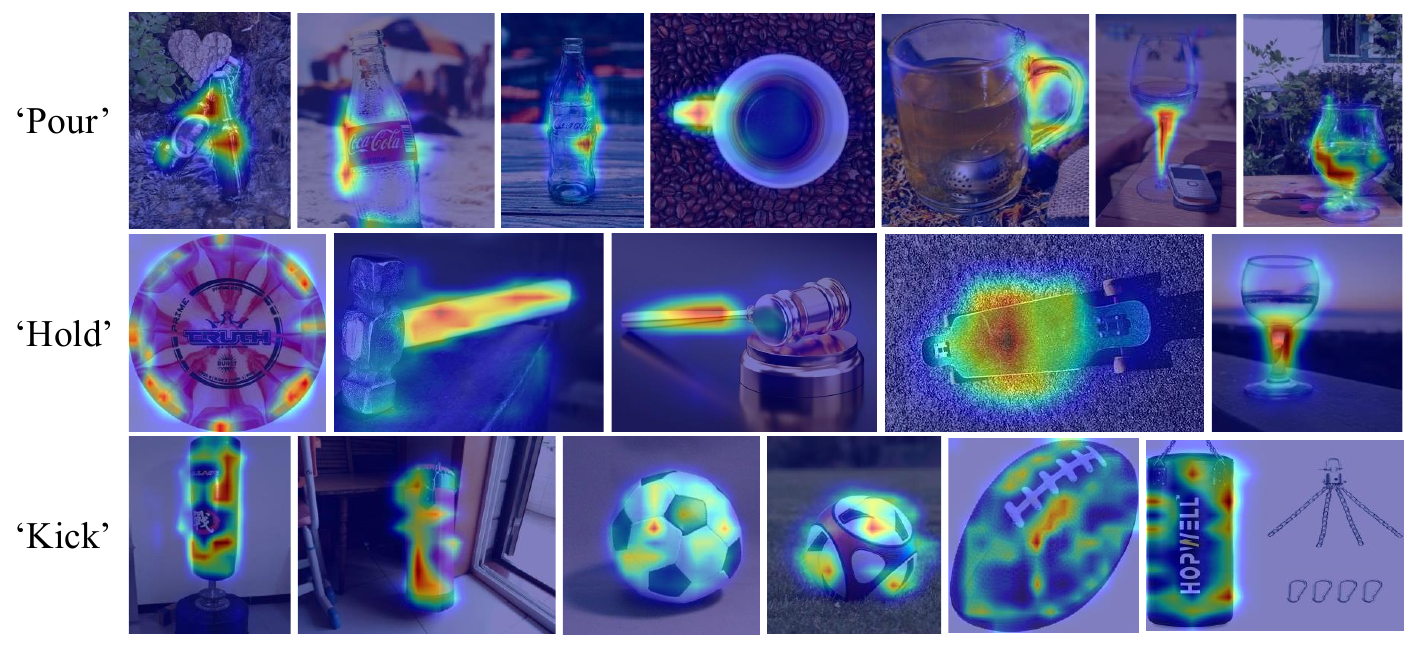}
    \caption{Additional Qualitative results of affordance grounding when interactions are unseen. The affordance grounding output alongside each interaction demonstrates the inference results for interactions that were not part of the training data. For example, in case of `pour', all the exocentric images related to `pour' were excluded during training, yet our model still exhibits fine grounding quality when inferring `pour' on `bottle', `cup', and `wine glass'.}
    \label{fig:fig_us_int}
\end{figure*}

With VLM text embedding, INTRA (Ours) successfully predicts affordance maps for novel interactions, as illustrated in our main paper. We designed and conducted experiments to analyze the generalization ability adopted by the VLM text encoder. During training, we excluded specific interaction classes and inferred affordance grounding on these excluded interaction classes to assess the grounding performance of the unlearned interactions. Fig.~\ref{fig:fig_us_int} is qualitative results of affordance grounding when the `hold', `pour', and `kick' classes are unseen during the training.

\subsubsection{Novel object and novel interaction.\label{nonni}}
\begin{figure*}[!p]
    \centering
    \includegraphics[width=\linewidth]{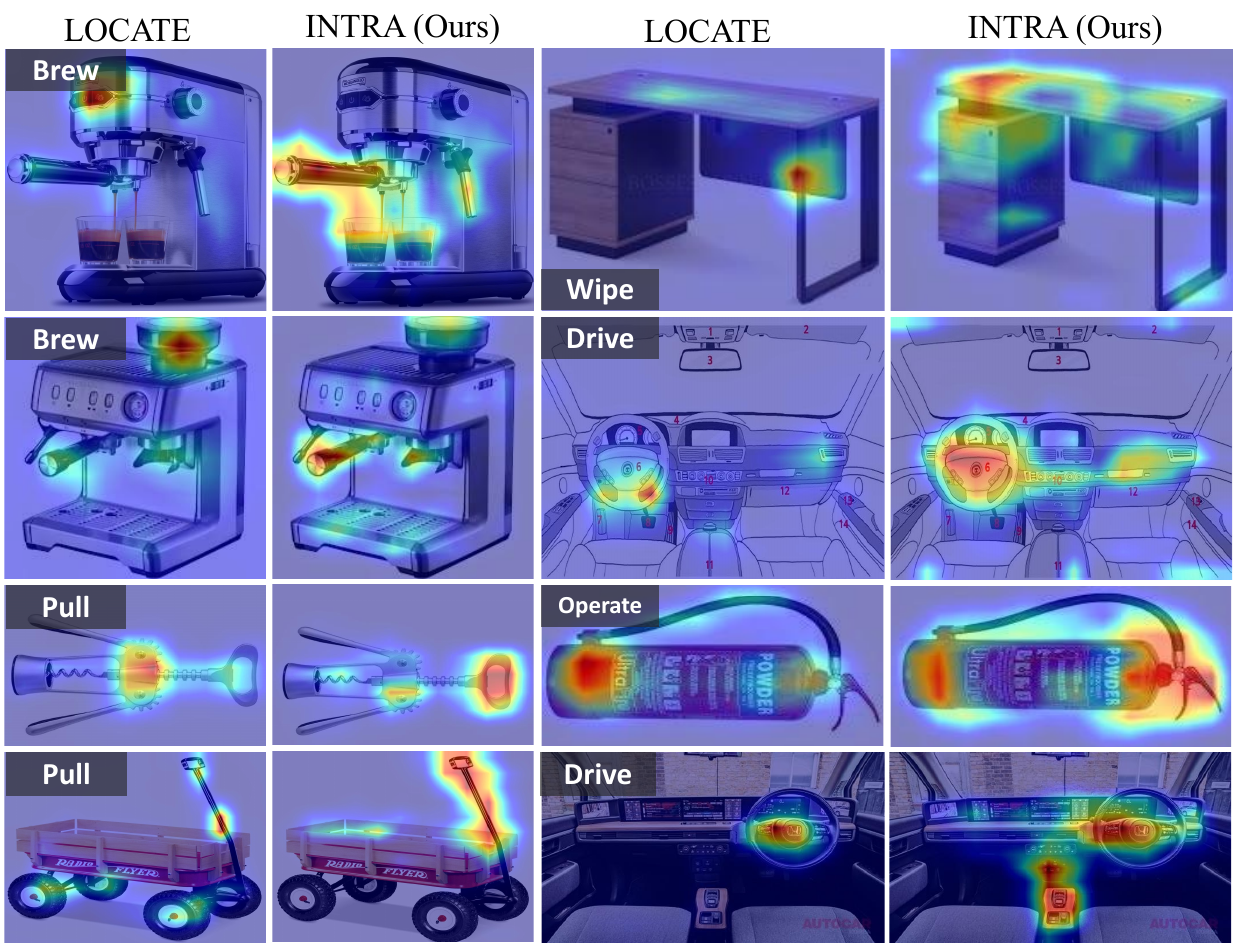}
    \caption{Additional qualitative results comparison between INTRA (Ours) and other baseline\cite{li2023locate} on affordance grounding when both the interaction and object are unseen during the training. Our approach accurately grounds affordances even when objects have many tractable parts, as observed in cases such as `coffee machine', `wine opener' or `car interior'. For instance, in the case of `brew', INTRA (Ours) captures the handle of the portafilter, while LOCATE~\cite{li2023locate} focuses on the bean container. Similarly, for the example of `wipe', INTRA grounds on the flat part of the desk, wheares LOCATE focuses on the desk leg.}
    \label{fig:fig_us_obj_int}
\end{figure*}

We conducted experiments to assess whether our model can infer affordance even in cases where both the interaction and object are novel. Since the baseline model can only infer predetermined interaction types, for comparison, we asked LLM to find the closest interaction of the novel interaction and inferred affordance grounding on the closest interaction in the baseline model. Specifically, `brew-pour', `wipe-brush with', `pull-hold', `drive-ride' were selected to substitute novel interaction in the baseline model. 
In Fig.~\ref{fig:fig_us_obj_int}, it is evident that our INTRA can accurately ground affordance even when both the interaction and object are unseen, and the novel objects have many tractable parts. For instance, Ours grounded all relevant parts of `Drive' in photos of car interior, whereas the baseline could not. Also, Ours accurately grounded all relevant parts that can be pulled in a wine opener and wagon. 

\subsubsection{Comparison on different scales.\label{different_scales}}
To investigate how varying affordance region scales impact the model, we follow \cite{li2023locate, luo2022learning, luo2022grounded}, dividing the test set into three subsets: `Big', `Medium' and `Small'. These subsets are determined by the ratio of the mask, with thresholds set at more than 0.1, between 0.03 and 0.1, and less than 0.03, respectively. INTRA (Ours) outperforms other baselines\cite{mai2020erasing, pan2021unveiling, gao2021ts, nagarajan2019grounded, luo2022grounded, luo2022learning, li2023locate} significantly in the `Small' subset, demonstrating its ability to generate finer heatmaps. Additionally, performance in the `Medium' and `Big' subsets is either comparable to or surpasses the baselines.

\section{Limitation}
While INTRA produces good grounding results even with novel interactions or objects, it finds it challenging to learn interactions in exocentric images where there is no contact between the object and the human. Although such interactions do not exist in AGD20K, it has been observed that classes without direct contact, such as `watch (clock)' or `direct (airplane),' are difficult to learn when training with the HOI datasets~\cite{chao:iccv2015, Pratt2020Swig, Wang_2021_ICCV, jiang2022bongard}. Thus, inferrring instructions that contain implicit meanings can be challenging. For example, grounding is possible for `turn on (air-conditioner),' but not for abstract instructions like `I'm cold.' Additionally, grounding may fail when the points of interaction are visually similar. For instance, when operating a microwave, it is difficult to identify the button with the affordance for `heat' among multiple buttons.

\end{document}